%% file: main.tex
\documentclass{article}

\usepackage[preprint,nonatbib]{neurips_2025}

\usepackage[utf8]{inputenc} 
\usepackage[T1]{fontenc}    
\usepackage{hyperref}       
\usepackage{url}            
\usepackage{booktabs}       
\usepackage{amsfonts}       
\usepackage{nicefrac}       
\usepackage{microtype}      
\usepackage{xcolor}         
\usepackage{graphicx}
\usepackage{algorithm}
\usepackage{algpseudocode}
\usepackage{multirow}
\usepackage{makecell}
\usepackage{hyperref}
\usepackage{enumitem}
\usepackage{amsmath}
\usepackage{cleveref}
\usepackage{wrapfig}
\usepackage{arydshln}
\usepackage{wrapfig}
\definecolor{citecolor}{HTML}{0071BC}
\definecolor{linkcolor}{HTML}{ED1C24}
\usepackage{xspace}
\makeatletter
\DeclareRobustCommand\onedot{\futurelet\@let@token\@onedot}
\def\@onedot{\ifx\@let@token.\else.\null\fi\xspace}
\def\eg{\emph{e.g}\onedot}

\makeatother

\title{A Self-supervised Motion Representation for Portrait Video Generation}

\author{%
  Qiyuan Zhang\textsuperscript{1}, 
    Chenyu Wu\textsuperscript{1}, 
    Wenzhang Sun\textsuperscript{2},
    Huaize Liu\textsuperscript{2,3,4},\\
    \textbf{Donglin Di}\textsuperscript{2},
    \textbf{Wei Chen}\textsuperscript{2},
    \textbf{Changqing Zou}\textsuperscript{1,3}
    \\
    \textsuperscript{1}Zhejiang University,
    \textsuperscript{2}Li Auto, 
    \textsuperscript{3}Zhejiang Lab,\\
    \textsuperscript{4}Hangzhou Institute for Advanced Study,University of Chinese Academy of Sciences
}

\begin{document}

\maketitle

\input{sections/0_abstract}

\input{sections/1_intro}

\input{sections/2_relatedwork}
\input{sections/3_method}

\input{sections/4_experiment}

\input{sections/6_conclusion}

\bibliographystyle{plain}
\bibliography{main}

\clearpage
\input{sections/7_supplement}
\clearpage
\input{sections/8_checklist}

\end{document}

%% file: sections/0_abstract.tex
\begin{abstract}

Recent advancements in portrait video generation have been noteworthy. However, existing methods rely heavily on human priors and pre-trained generative models, Motion representations based on human priors may introduce unrealistic motion, while methods relying on pre-trained generative models often suffer from inefficient inference. To address these challenges, we propose \textbf{Se}mantic Latent \textbf{Mo}tion (\textbf{SeMo}), a compact and expressive motion representation. Leveraging this representation, our approach achieve both high-quality visual results and efficient inference. SeMo follows an effective three-step framework: Abstraction, Reasoning, and Generation. First, in the Abstraction step, we use a carefully designed Masked Motion Encoder, which leverages a self-supervised learning paradigm to compress the subject's motion state into a compact and abstract latent motion (1D token). Second, in the Reasoning step, we efficiently generate motion sequences based on the driving audio signal. Finally, in the Generation step, the motion dynamics serve as conditional information to guide the motion decoder in synthesizing realistic transitions from reference frame to target video. Thanks to the compact and expressive nature of Semantic Latent Motion, our method achieves efficient motion representation and high-quality video generation. User studies demonstrate that our approach surpasses state-of-the-art models with an 81\% win rate in realism. Extensive experiments further highlight its strong compression capability, reconstruction quality, and generative potential. 

\end{abstract}

%% file: sections/1_intro.tex
\section{Introduction}
\label{sec:intro}

Portrait video generation has gained significant attention due to its numerous applications in gaming, film, and education. With the development of AI generation technologies, there is an increasing demand for more realistic animation effects and faster generation speeds.

\begin{figure}[t]\centering
\includegraphics[width=0.99\linewidth]{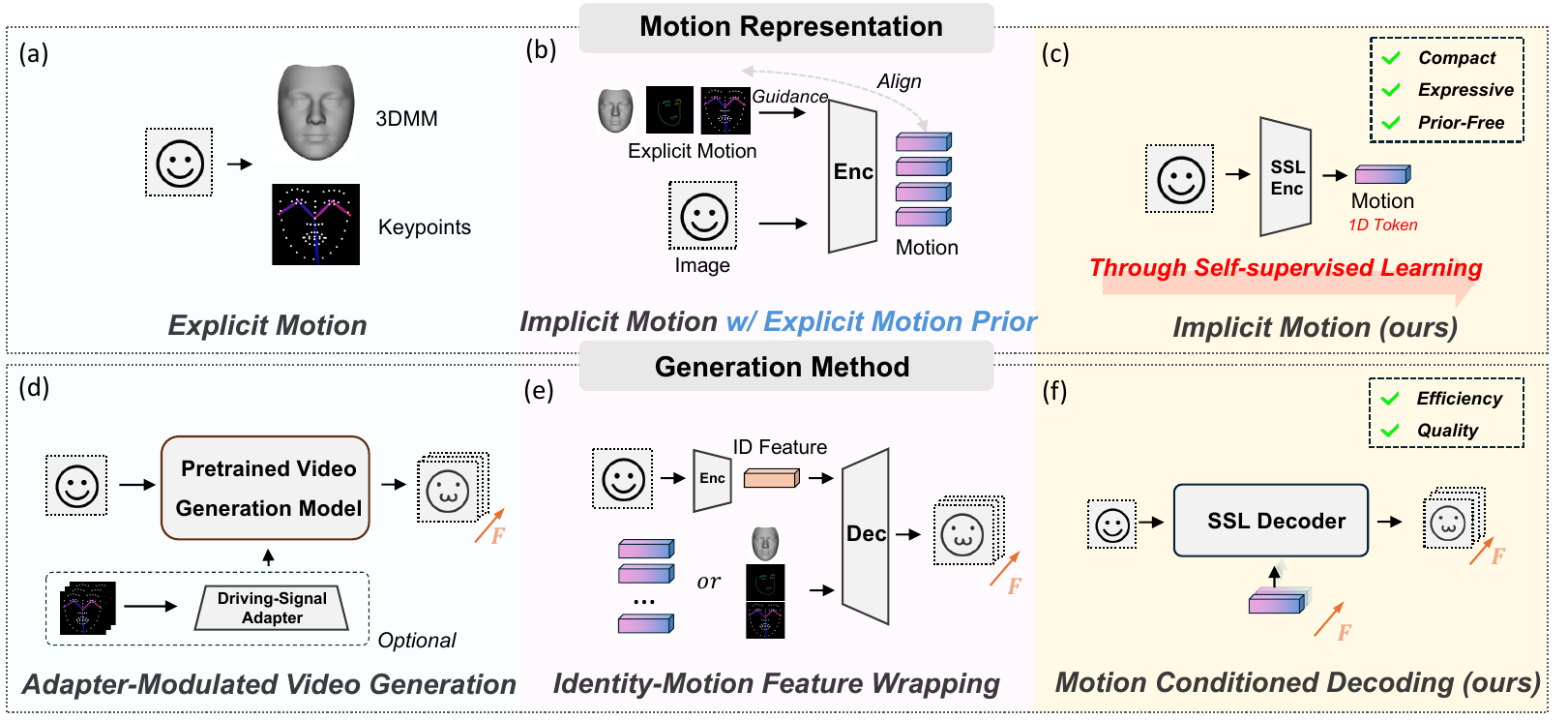}
\vspace{-4mm}
\caption{\textbf{Comparison of Motion Representation and Generation Methods.}
Motion representation:
(a) Explicit motion representation (\eg, 3DMM, facial landmarks).
(b) Implicit motion representation guided by explicit priors (\eg, implicit keypoints).
(c) Our self-supervised implicit representation compresses motion into a compact 1D token without pretrained models or prior knowledge.
Video generation approaches:
(d) Adapter-modulated generation fine-tunes a pretrained video model via driving-signal adapters.
(e) Identity-motion feature wrapping fuses identity features with target motion.
(f) Our method leverages reference and target motion as conditions for efficient generation.}

\label{fig:fig1}
\vspace{-4mm}
\end{figure}

Existing approaches mainly follow two paths. In the first path (\cref{fig:fig1}e), methods directly utilize explicit motion representations, or employ explicit priors to learn an implicit motion representation; these motion representations are then fused with identity features to generate video frames. In the second path (\cref{fig:fig1}d), pre-trained video generation models are used, utilizing their powerful generative capacity and conditional cross-attention mechanisms to learn the mapping from the driving signal to video frames~\cite{cui2024hallo2,echomimic2024,aniportrait2024}. Many methods combine both of these paths. However, in the first path, human prior knowledge relies on domain-specific heuristics, often failing to capture detailed motions, which results in human videos that lack naturalness and coherence. Moreover, this framework is not adaptable to multiple scenarios (\eg, 3DMM is only applicable to facial scenarios and cannot be generalized to other types of scene). In the second path, the computationally intensive spatial-temporal attention operations - with their quadratic scaling in memory and computation - combined with multistep denoising in diffusion models create significant bottlenecks for video generation efficiency.


We observe that these motion representations (3DMM, pose, landmarks) all express a high-level semantic:`` the positions and attributes of meaningful parts.'' They are all abstractions of human knowledge about facial dynamics and are limited by design biases. For instance, we place landmarks on faces because we believe that changes in the positions of the noise, eyes, and mouth corners are crucial for facial dynamics. Similarly, we use skeletons because we believe that skeletal structures are important for body movements. However, these representations are limited by human abstracting capabilities and cannot capture detailed motion. We investigate the fundamental question: \textit{How can we derive a motion representation that effectively captures fine-grained facial dynamics without relying on prior knowledge?} To address this, we first present our motivation from two key perspectives.


\textbf{1) Video motion in a highly compact and semantic space.}  Prior works~\cite{tian2024reducio,yu2024efficient,wang2024vidtwin} have demonstrated that the content between frames in general videos contains substantial redundancy, video motion can be compressed into a very small latent space. For portrait videos in particular, the motion can be represented even more compactly, as the motion range of human subjects is inherently limited and the background typically remains static. Previous approaches leverage human-crafted semantic priors (such as 3DMM or landmarks) to represent motion through domain knowledge, and have validated the effectiveness of this strategy. This suggests that, by surpassing the abstraction capacity and compression limitations of human-designed priors, it may be possible to encode portrait video motion into an even more compact and semantic latent space.

\textbf{2) Self-supervised representations as guidance.} Current generative models~\cite{cho2024sora,li2024survey,HumanVideo2018,everydance2019,posevideo2018,Latte2024,ControlNet,guo2023animatediff} predominantly rely on human-abstracted conditions like text prompts or landmarks. While effective, these representations are inherently limited by human abstraction capabilities and domain-specific biases, often failing to capture subtle motion details. Recent works like RCG~\cite{rcg2024} and StoryDiffusion\cite{zhou2024storydiffusion} have demonstrated that self-supervised latent features can serve as superior generation conditions compared to manual priors. Building on this insight, we propose using learned semantic latent motion as conditional signals for video generation.


Motivated by these insights, we propose a universal motion representation framework paired with a simple yet effective video generation pipeline. Our approach comprises two key stages: In the first stage, we train a Motion Autoencoder in which the encoder compresses each video frame into a compact latent motion representation (1D token) using a random masking strategy. This encourages the model to abstract relevant semantic motion while discarding irrelevant details. The rectified flow-based decoder then leverages its powerful generative capabilities to fill in visual details, conditioned on the reference image and reference motion. In the second stage, we train a standard DiT model to directly generate motion sequences from audio and reference motions. Thanks to the highly compact nature of our motion representation, this process is extremely efficient. Our motion extraction and video generation pipeline is both simple and straightforward to generalize to other scenarios, requiring neither pretrained models nor hand-crafted human priors. Benefiting from the strong descriptive power of our semantic latent motion representation, our method produces highly natural and vivid animations, surpassing state-of-the-art approaches with an FID score of 26.8 (compared to a baseline of 27.4) and achieving an 81\% win rate in realism according to user studies. In summary, our contributions to the community include:

\begin{enumerate}[topsep=3.5pt,itemsep=3pt,leftmargin=20pt]
\item We present a new representation framework that fuses the semantic descriptive capability of self-supervised representations with the fine-grained generative fidelity of a rectified-flow model, offering a new perspective on dynamic representation.

\item We show that the proposed representation encodes rich, complex motion cues in an compact latent space, leading to better reconstruction quality in video reconstruction tasks.

\item We validate the effectiveness of our approach in audio-driven portrait video generation. Owing to the self-supervised nature of SeMo, which eliminates the need for pretrained models or manually-designed priors, our method adapts well to a wide range of challenging scenarios while maintaining strong performance.
\end{enumerate}

%% file: sections/2_relatedwork.tex
\vspace{-2mm}
\section{Related Works}
\label{sec:related works}




\paragraph{Motion-Based Video Codec.}
Motion-driven video synthesis techniques typically generate target frames by combining reference frames with motion representations. FOMM~\cite{fomm2020} established a baseline through sparse keypoint detection and affine transformation estimation to derive dense motion fields. LIA~\cite{lia2024} advanced this paradigm by decomposing latent vectors and employing optical flow-based warping operations. MRAA~\cite{mraa2021} extended the framework by separately predicting foreground region motions and background displacements, subsequently rendering through pixel-wise flow manipulation. IMF~\cite{imf2024} introduced a fully self-supervised approach that encodes reference and target frames into compact latent tokens. The work most closely related to ours is IMF~\cite{imf2024}.~\cref{suppsec:novelty} illustrates the key differences between our approach and IMF.

\paragraph{Portrait Video Generation.}
The portrait generation task has garnered significant attention in recent years due to its broad applications and has witnessed rapid advancements. Wave2Lip~\cite{wav2lip2020} pioneered the use of GAN networks combined with lower-face masking to directly synthesize video frames from reference images and audio inputs. Subsequent works including VividTalk~\cite{vividtalk2023}, SadTalker~\cite{sadtalker2023}, and Moda~\cite{MODA2023} have leveraged 3D morphable priors coupled with neural rendering techniques to generate target video frames. Notably, VASA-1~\cite{vasa2024} and Gaia~\cite{gaia2024} introduced hybrid architectures that encode both 3D prior knowledge and appearance features into a latent space, establishing a more universal and compact motion representation framework. Next, generative models are employed to produce motion sequences conditioned on audio inputs~\cite{emo2024, liu2025moee, sun2024uniavatar, zhang2024musetalk}. Recent innovations like EMO~\cite{emo2024} and Hallo2~\cite{cui2024hallo2} have further enhanced visual quality through the integration of pre-trained video generation foundation models with specialized reference networks. AniTalker~\cite{anitalker2024} presents a two-stage self-supervised learning paradigm. 

\paragraph{Self-supervised Representation as Guidance.}
A large amount of self-supervised learning work ~\cite{mae2022,sapiens2024,videomae2022,Girdhar2023OmniMAE,qian2021spatiotemporal,yang2020video,wang2023VideoMAE} has utilized encoded features for various downstream visual understanding tasks. Self-supervised representations have been proven to possess rich semantics and can serve as powerful guidance signals. In RCG ~\cite{rcg2024}, the quality of image generation was significantly improved by introducing features obtained through MoCo’s~\cite{moco2021} self-supervised learning framework. Meanwhile, Story Diffusion~\cite{zhou2024storydiffusion} obtains a continuous guidance signal by performing linear interpolation in the latent space of CLIP features, thereby controlling the video generation model to produce more coherent content. REPA~\cite{yu2025repa} and VA-VAE~\cite{yao2025vavae} demonstrate the ability to improve generation and reconstruction by aligning latent features with strong self-supervised visual representations.
\vspace{-2mm}

\begin{figure*}\centering
\includegraphics[width=1.0\linewidth]{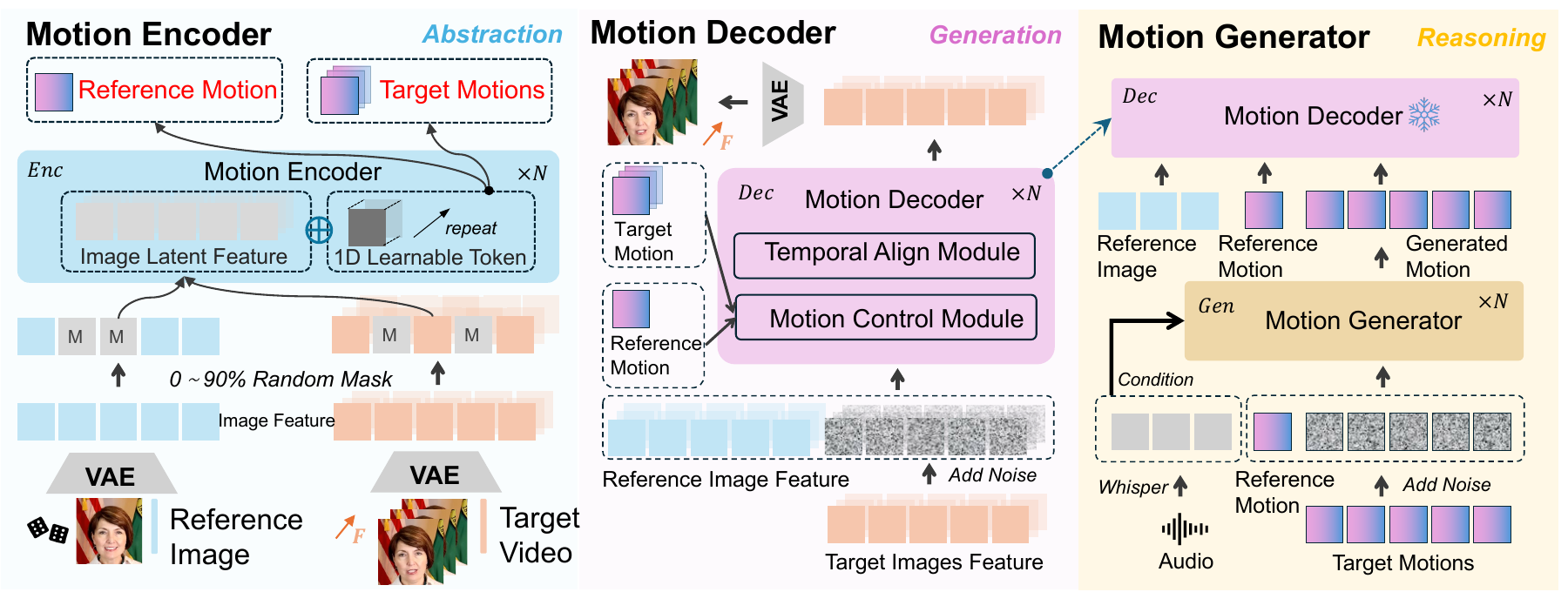}
\vspace{-2mm}
\caption{\textbf{Overview of SeMo.} The proposed framework follows a three-step process (abstraction, reasoning, generation) implemented in two training stages. In the first stage, we train a Motion AutoEncoder for abstraction and generation. During the abstraction step, image features are subjected to random masking and compression to obtain a compact and abstract latent motion. In the generation step, latent motion is used as a condition to guide the motion decoder in generating the target video from the reference image. In the second stage, we train a Motion Generator for reasoning, which generates motion sequences from audio and reference motions. }
\label{fig:pipeline}
\vspace{-4mm}
\end{figure*}


%% file: sections/3_method.tex
\newpage
\section{Method}
\label{sec:method}
\vspace{-2mm}
Given a reference image and its corresponding audio, our goal is to generate a dynamic video with synchronized motion and sound. To achieve this, we propose a two-stage self-supervised learning framework. In the first stage, the objective is to extract a compact, high-level semantic motion $m \in \mathbb{R}^{l\times d}$ for each frame. We first use learnable tokens and random masking strategy to compress the motion into an abstract representation, and then utilize the reference image, reference motion and target motion to reconstruct the videos. In the second stage, we train a Motion Generator and the Motion AutoEncoder is frozen. For each frame, we first use the motion encoder to extract the motion and then generate motion sequence based on the audio and reference motion. The details of these two stages will be described below.

\subsection{Motion AutoEncoder}
Our core motivation is to leverage self-supervised learning to extract the most salient motion information, and then utilize the powerful distribution modeling ability of rectified flow to progressively complete the high-frequency details conditioned on the motion information in a coarse-to-fine manner.

\noindent \textbf{Motion Encoder.}
\label{sec:encode}
Our motion encoder design is motivated by the observation that facial movements - akin to 3D Morphable Models (3DMM) and keypoint trajectories - can be effectively represented in a highly compact space. We propose a new Masked Motion Encoder architecture that integrates dual strategies of random masking and latent space compression to derive semantically rich latent motion.
Specifically, we first compress input frames $I \in \mathbb{R}^{H\times W\times C}$  into latent representations $X \in \mathbb{R}^{h\times w\times c}$ using a pre-trained stable diffusion VAE encoder \cite{sd2022}. Following ViT~\cite{vit2021}, we embed each latent into non-overlapping patches, and apply random masking with ratios uniformly sampled from 0\% to 90\%. Once the masking is complete, we discard the masked tokens and retain the unmasked tokens, arranging them into a sequence in their original order. The masking ratios are carefully designed: low mask ratios encourage the model to retain more complete information, while high mask ratios are intended to strengthen the model’s ability to ignore fine-grained details and focus on preserving the most salient motion information. As shown in~\cref{fig:userstudy}, applying a masking strategy can greatly enhance the model’s generalization ability by ignoring unnecessary information. The mask strategy can enhance the model’s generalization ability~\cref{suppfig:mask2} and also accelerate convergence during the second-stage training~\cref{fig:mask}. More analysis is provided in~\cref{sec:ablation_mask}.

TikTok \cite{tiktok2024} has demonstrated that image features can be represented as 1D sequence representations. Therefore, we designed a 1D learnable token $y \in \mathbb{R}^{l\times d}$ where $l$ denotes the number of tokens and $d$ denotes the features dimension. we concatenate the learnable token with the image features and process them through ViT layers. Ultimately, latent motion $M$ is output at the position corresponding to the learnable tokens. Here, the number of motion tokens and the token dimension size are consistent with the learnable tokens. Each image is encoded independently; we refer to the image features and motion associated with the reference image as $x_{\text{ref}}$,$m_{\text{ref}}$, and those associated with the target image as $x_{\text{tar}}$,$m_{\text{tar}}$. The encoding process can be formulated as follows: $m_{\text{ref}} = \operatorname{Enc}_{\theta}(x_{\text{ref}})$ , $m_{\text{tar}}= \operatorname{Enc}_\theta(x_{\text{tar}})$, where $\theta$ is the parameter of the motion encoder.

\begin{figure}[t]
\begin{center}
\vspace{-4mm}
\begin{minipage}[c]{0.38\textwidth}
    \includegraphics[width=\textwidth]{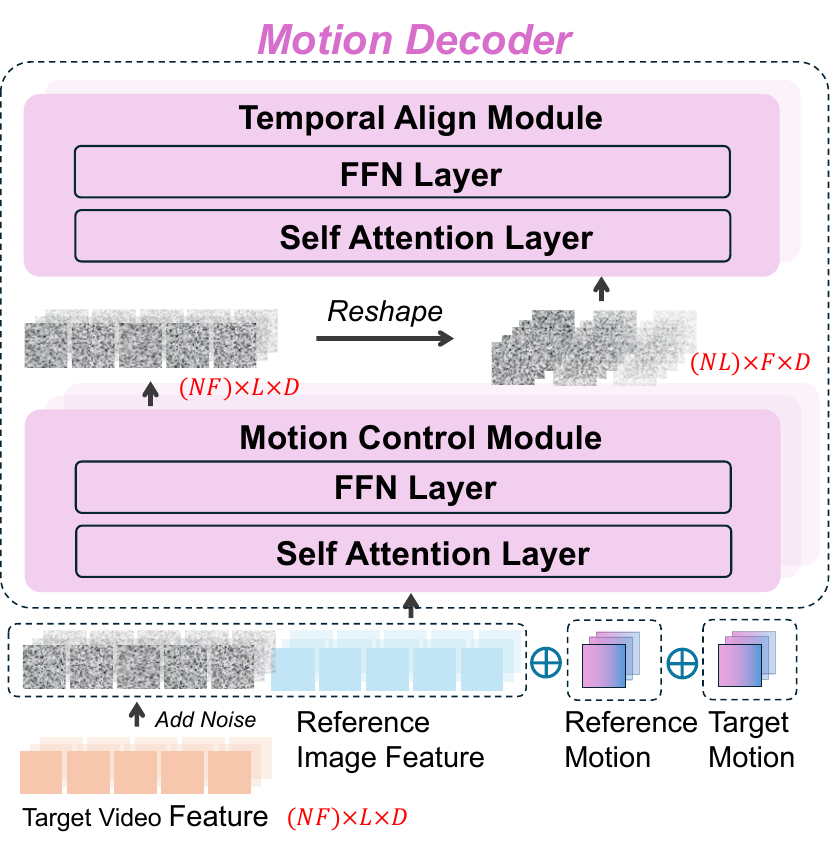}
\end{minipage}
~
\hspace{2mm}
\begin{minipage}[c]{0.42\textwidth}
\vspace{-2mm}
    \caption{\textbf{Motion Decoder.} The Motion Control Module uses the reference motion $m_{\text{ref}}$ and target motion $m_{\text{tar}}$ as control signals. The goal of this module is to transfer from the reference image features $x_{\text{ref}}$ to the target image features $x_{\text{tar}}$. After reshaping, the target image features $x_{\text{tar}}$ are processed with self-attention along the temporal dimension, enabling alignment of high-frequency information.}\label{fig:decoder}
\end{minipage}
\end{center}
\vspace{-8mm}
\end{figure}

\noindent\textbf{Motion Decoder.}
At the decoding stage, our objective is to use the reference image for appearance information, with motion serving as the semantic condition, guiding a generative model to fill in missing details. 
Our decoder consists of two modules: the Motion Control Module and the Temporal Align Module (MCM and TAM for short). In the Motion Control Module, we concatenate the reference image features $x_{\text{ref}}$, the noised target image features $\bar{x}_{\text{tar}}$, the reference motion $m_{\text{ref}}$, and the target motion $m_{\text{tar}}$, then process them through a self-attention layer followed by a feed-forward network (FFN). The noising process follows the schedule in rectified flow, which can be formulated as: $\bar{x} = (1-t)x + t \epsilon$, where $t \in [0, 1]$ is the noise level, and $\epsilon$ denotes Gaussian Noise sampled from standard normal distribution. $z_{\text{tar}} = \operatorname{MCM}_{\psi_1}(x_{\text{ref}}, \bar x_{\text{tar}}, m_{\text{ref}}, m_{\text{tar}})$.
Since denoising process in Motion Control Module is performed on each target frame, there could be inconsistencies in the high-frequency details across target frames. Therefore, in the Temporal Align Module, we concatenate the target image features $\{ z_{\text{tar}}^f\}_{f=1}^k$ along the temporal dimension and apply self-attention to align their high-frequency information, where $k$ is the number of frames. $\{x_{\text{tar}}^f\}_{f=1}^k = \operatorname{TAM}_{\psi_2}(\{z_{\text{tar}}^f\}_{f=1}^k)$. In all, the decoding process can be formulated as follows: $\{x_{\text{tar}}^f\}_{f=1}^k = \operatorname{Dec}_{\psi}(x_{\text{ref}}, \{\bar{x}_{\text{tar}}^f\}_{f=1}^k , m_{\text{ref}} , \{m_{\text{tar}}^f\}_{f=1}^k)$, where $\psi$ is the parameter of the motion decoder.

\vspace{-2mm}
\subsection{Motion Generator}
 Let $p$ be the number of previous frames input.
 In this module, our objective is to train a motion generator that can generate a corresponding motion sequence based on a previous motion $\{m_{\text{pre}}^f\}_{f=1}^{p}$, noised target motion $\{\bar{m}_{\text{tar}}^f\}_{f=1}^k$ and an audio sequence $\{a^f\}_{f=1}^k$.
 Note that $\{m_{\text{pre}}^f\}_{f=1}^{p}$ and $\{m_{\text{tar}}^f\}_{f=1}^{k}$ is extracted by motion encoder from corresponding image features without mask and the audio sequence $\{a^f\}_{f=1}^k$ is encoded by a pretrained model Whisper~\cite{whisper2022}. Specifically, we concatenate the noised target motions $\{{\bar{m}}_{\text{tar}}^f\}_{f=1}^k$ with previous motion $\{m_{\text{pre}}^f\}_{f=1}^p$ for self-attention modeling, while cross-attention layers establish motion-audio alignment. As shown in~\cref{fig:mask}, the masking strategy improve the distribution of motion latents, which accelerates the convergence as well as the performance of the second-stage training. 
 The motion generation process can be formulated as follows: $\{m_{\text{tar}}^f\}_{f=1}^k = \operatorname{Gen}_{\phi}( \{\bar{m}_{\text{tar}}^f\}_{f=1}^k , \{m_{\text{pre}}^f\}_{f=1}^p , \{a^f\}_{f=1}^k)$, where $\phi$ is the parameter of the motion generator.




\vspace{-2mm}
\subsection{Training Objective}
In the first stage, the goal of the motion autoencoder is to reconstruct the target video based on the reference image and corresponding motion. In the second stage, the model is optimized to generate motion sequence based on audio condition. Both stage utilize rectified flow as the generative framework. Let $v_{\psi}$ and $u_{\phi}$ denote the velocity respectively, we optimize the distance between the predicted velocity $v_{\theta}$ and the true velocity $v$.

\vspace{-3mm}
\begin{equation} \label{loss1}
\min_{\theta,\psi}\mathcal{L}_{\text{first stage}}=\min_{\theta,\psi}\mathbb{E}_{t, x_{\text{ref}},x_{\text{tar}}, \epsilon}\left[\left\|(x_{\text{tar}}-\epsilon)-v_{\psi}\left(x_{\text{ref}},\bar x_{\text{tar}}, \operatorname{Enc}_{\theta}(x_{\text{ref}}),\operatorname{Enc}_{\theta}(x_{\text{tar}}), t\right)\right\|^{2}\right]
\end{equation}
\vspace{-3mm}
\begin{equation} \label{loss2}
\min_{\phi}\mathcal{L}_{\text{second stage}} =\min_{\phi}\mathbb{E}_{{t}, a, m_{\text{tar}}, m_{\text{pre}}, \epsilon}\left[\left\|(m_{\text{tar}}-\epsilon)-u_{\phi}\left(a,\bar m_{\text{tar}}, m_{\text{pre}}, t\right)\right\|^{2}\right]
\end{equation}






%% file: sections/4_experiment.tex
\section{Experiment}
\label{sec:exp}
\vspace{-2mm}
Our experiments are mainly divided into three parts. First, we verify the effectiveness of the representations on the video reconstruction task~\cref{sec:Reconstruction}. Then, we evaluate the applicability of the representations to downstream tasks by testing on audio-driven portrait video generation~\cref{sec:generation}. Finally, we conduct extensive ablation studies to demonstrate the necessity of our experimental strategies~\cref{sec:ablation}.

\vspace{-2mm}
\subsection{Experimental Settings}

\begin{figure*}[t]\centering
\includegraphics[width=1.0\linewidth]{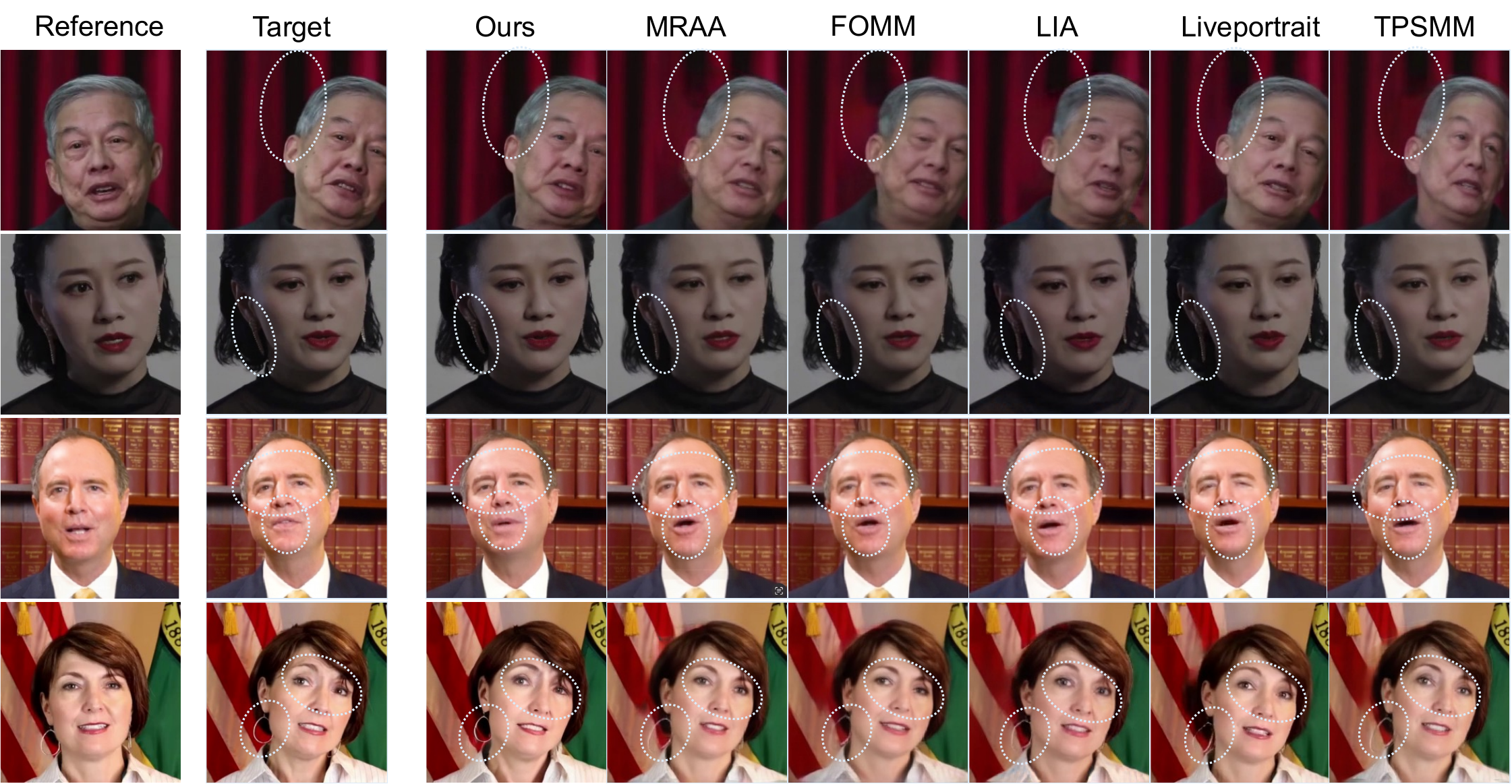}

\caption{\textbf{Results of video reconstruction.} We show video reconstruction results on datasets DH-FaceVid-1K and HDTF.}
\label{fig:reconstruction}
\vspace{-2mm}
\end{figure*}

\begin{figure}[t]\centering
\includegraphics[width=1.0\linewidth]{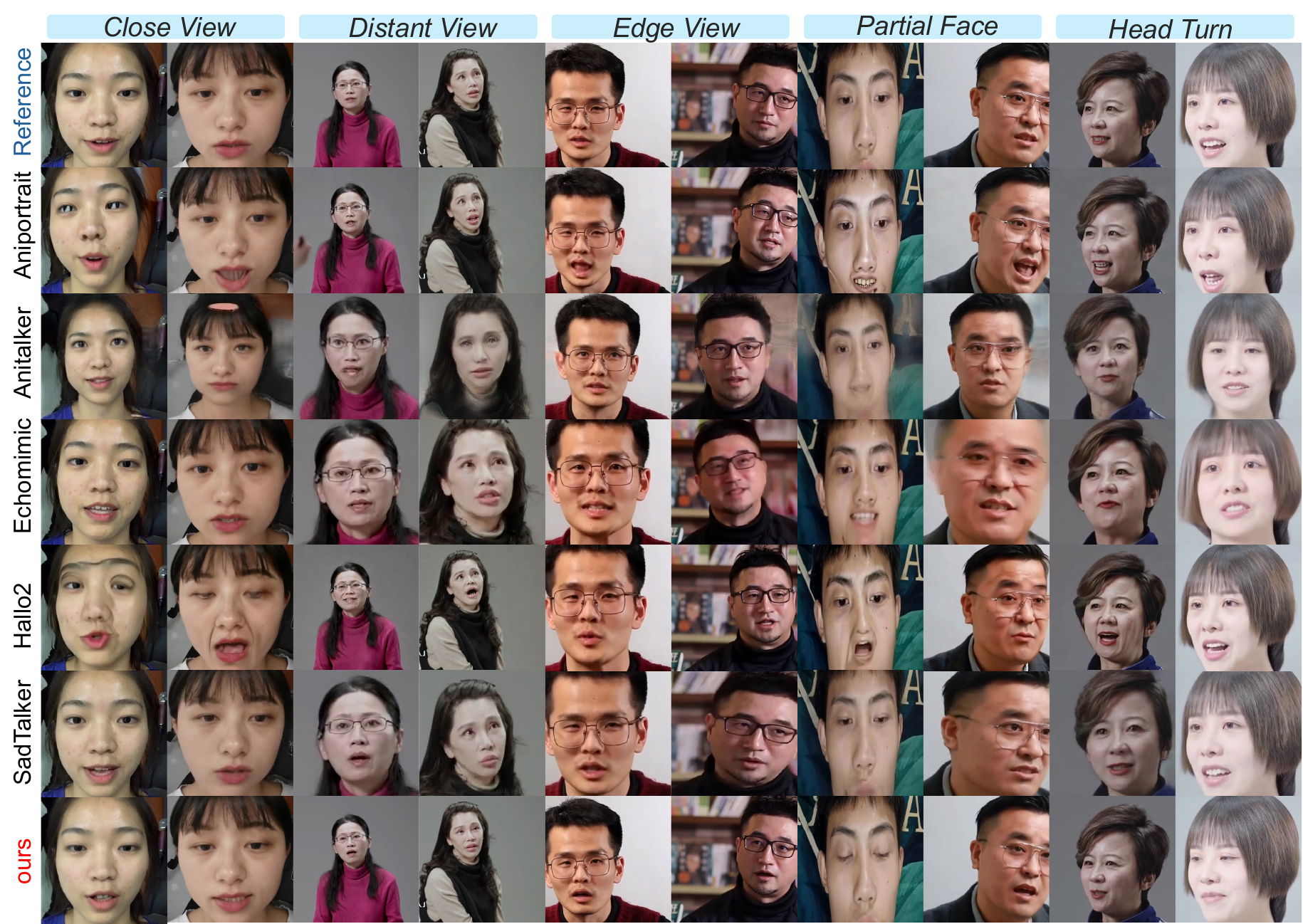}
\caption{\textbf{Result of Audio-driven Video Generation.} Our method adapts to various challenging scenarios. Zoom in for a better view. (a) Close view: the subject is very close to the camera; (b) Distant view: the subject is far from the camera, nearly half-body is shown; (c) Edge view: the face appears at the edge of the frame; (d) Partial face: only part of the face is visible; (e) Head turn: the subject turns their head.}
\label{fig:generation}
\vspace{-4mm}
\end{figure}



\begin{table*}[t]
\begin{center}
\caption{Motion-based video reconstruction comparison of various methods on the HDTF, DH-FaceVid-1K, and MEAD dataset. Red indicates the best result, blue the second best, and green the third best.}
\label{tab:reconstruction}
\vspace{-2mm}
\resizebox{1\textwidth}{!}{
	\begin{tabular}{lccccc|ccccc|ccccc}
        \toprule
        \multirow{2}{*}{\textbf{Method}} & \multicolumn{5}{c}{\textbf{HDTF}} & \multicolumn{5}{c}{\textbf{DH-FaceVid-1K}} & \multicolumn{5}{c}{\textbf{MEAD}}                                                
        \\ 
        \cmidrule(lr){2-6} \cmidrule(lr){7-11} \cmidrule(lr){12-16}
        & \textbf{FID$\downarrow$} & \textbf{FVD$\downarrow$} & \textbf{LPIPS$\downarrow$} & \textbf{PSNR$\uparrow$} & \textbf{SSIM$\uparrow$}  &
        \textbf{FID$\downarrow$} & \textbf{FVD$\downarrow$} & \textbf{LPIPS$\downarrow$} & \textbf{PSNR$\uparrow$}  & \textbf{SSIM$\uparrow$}  &
        \textbf{FID$\downarrow$} & \textbf{FVD$\downarrow$} & \textbf{LPIPS$\downarrow$} & \textbf{PSNR$\uparrow$}  & \textbf{SSIM$\uparrow$} 
        \\ \hline

        FOMM\cite{fomm2020}  & 24.690 & 188.797 & 0.064 & 25.572 & 0.816  & 30.052 & 236.578 & 0.061 & 25.542 & 0.844 & 41.541 & 166.950 & 0.040 & 31.131 & 0.917            \\
        
        MRAA\cite{mraa2021}  & \textcolor{green}{23.004} & 155.53 & \textcolor{green}{0.046} & \textcolor{blue}{29.411} & \textcolor{green}{0.891} & \textcolor{blue}{22.258} & 216.034 & \textcolor{green}{0.044} & \textcolor{blue}{30.295} & \textcolor{green}{0.913} & \textcolor{green}{35.378} & 167.186 & \textcolor{green}{0.030} & \textcolor{green}{34.245} & \textcolor{blue}{0.950}           \\
        
        TPSMM\cite{zhao2022thin}  & 26.556 & \textcolor{green}{138.88} & \textcolor{red}{0.041} & \textcolor{red}{29.51} & \textcolor{blue}{0.901} & \textcolor{green}{24.697} & \textcolor{green}{215.362} & \textcolor{blue}{0.042} & \textcolor{green}{29.625} & \textcolor{red}{0.917} & \textcolor{blue}{34.358} & \textcolor{green}{136.651} & \textcolor{blue}{0.028} & \textcolor{blue}34.805 & \textcolor{red}{0.956}  \\
        
        LIA\cite{lia2024}  & 26.366 & 186.307 & 0.067 & 26.624 & 0.840 & 33.409 & 238.101 & 0.067 & 26.876 & 0.861 & 47.152 & 170.724 & 0.048 & 31.124 & 0.921            \\
        
        Liveportrait\cite{guo2024liveportrait}  & \textcolor{blue}{22.469}  & \textcolor{blue}{137.935} & 0.068 & 24.810 & 0.802 & 26.502 & \textcolor{blue}{170.360} & 0.059 & 25.681 & 0.839 & 56.700 & \textcolor{blue}{88.304} & 0.038 & 31.661 & 0.923    \\
        
        SeMo~(Ours)  & \textcolor{red}{21.431} & \textcolor{red}{136.738} & \textcolor{blue}{0.043} & \textcolor{green}{29.12} & \textcolor{red}{0.912} & \textcolor{red}{22.045} & \textcolor{red}{112.166} & \textcolor{red}{0.036} & \textcolor{red}{32.132} & \textcolor{blue}{0.914} & \textcolor{red}{23.546} & \textcolor{red}{70.975} & \textcolor{red}{0.019} & \textcolor{red}{36.233} & \textcolor{green}{0.949}         

        \\ \bottomrule

        \end{tabular}
        \vspace{-5mm}
    }
\end{center}
\vspace{-10mm}
\end{table*}

\textbf{Datasets.}
For training the Motion AutoEncoder, we utilized three datasets: DH-FaceVid-1K~\cite{facevid2024}, HDTF~\cite{hdtf2021}, and MEAD~\cite{kaisiyuan2020mead}. Videos were resized and center-cropped to 256×256 resolution, excluding those failing face detection. Training samples were equally drawn from all three datasets.
For training the Motion Generator, we employed two datasets: HDTF and DH-FaceVid-1K. We removed samples with excessively noisy or unclear audio and encoded the audio using the pre-trained model Whisper \cite{whisper2022}.

\noindent \textbf{Implementation Details.}
All two-stage experiments were trained on 8 NVIDIA A800 GPUs. We set the dimension of the semantic latent motion to 1 token with 512 channels throughout all experiments to balance reconstruction accuracy and generation quality. In the first stage, the motion autoencoder generates a sequence of $k=16$ frames in a single forward pass. In the second stage, the motion generator also produces $k=16$ motion tokens  per forward pass aided with $p=8$ previous motion tokens. All models are trained under similar settings: a base learning rate of $10^{-4}$ per 64 batch size and the AdamW optimizer with $\beta_1=0.9$, $\beta_2=0.95$, $\text{decay}=0.05$.

\noindent \textbf{Evaluation Metrics.}
We evaluate our experiments very rigorously using various metrics. In the video reconstruction phase, we use FID~\cite{heusel2017gans}, FVD~\cite{wang2018videotovideosynthesis}, LPIPS~\cite{zhang2018lpips}, SSIM~\cite{wang2004image} and PSNR. In the audio-driven portrait video generation phase, we use FID, FVD, Sync-C and Sync-D.

\noindent \textbf{Baseline.}
In the motion-based video reconstruction phase, we compared
MRAA \cite{mraa2021}, FOMM \cite{fomm2020}, Lia \cite{lia2024}, Liveportrait \cite{guo2024liveportrait}, TPSMM \cite{zhao2022thin}. In the portrait video generation phase, we compared Aniportrait \cite{aniportrait2024}, Anitalker \cite{anitalker2024}, Echomimic \cite{echomimic2024}, Hallo2 \cite{cui2024hallo2}, SadTalker \cite{sadtalker2023}.

\subsection{Comparison}
\subsubsection{Motion-based Video Reconstruction.}
\label{sec:Reconstruction}

Motion-based video reconstruction refers to an algorithm that generates target frames based on the reference image and motion. In the experiment, \cref{tab:reconstruction} we directly applied their pretrained weights in the face reconstruction task. We observed that our method outperformed theirs in terms of metrics. Additionally, Since the training task involves reconstructing entire videos through self-supervised learning rather than focusing solely on the face, in~\cref{fig:reconstruction},we observe that the model is able to capture more detailed and complex motions, such as those of earrings and hairstyles. Furthermore, thanks to the powerful generative capability of the motion decoder, the model can automatically complete missing details caused by head movements. 

Methods based on explicit representations~\cite{fomm2020,mraa2021,zhao2022thin} perform poorly, especially in cases involving large or rapid motions. They often fail to capture fine-grained, high-frequency changes. Moreover, these explicit representations are typically designed for the face only, neglecting global motion information, which further limits their performance and robustness. As a result, such methods tend to generalize poorly and cannot be effectively transferred to other scenarios. We further provide extensive insights on the mask~\cref{suppsec:mask} and scene transferability~\cref{suppsec:general}.


\begin{figure}[t]\centering
\includegraphics[width=1.0\linewidth]{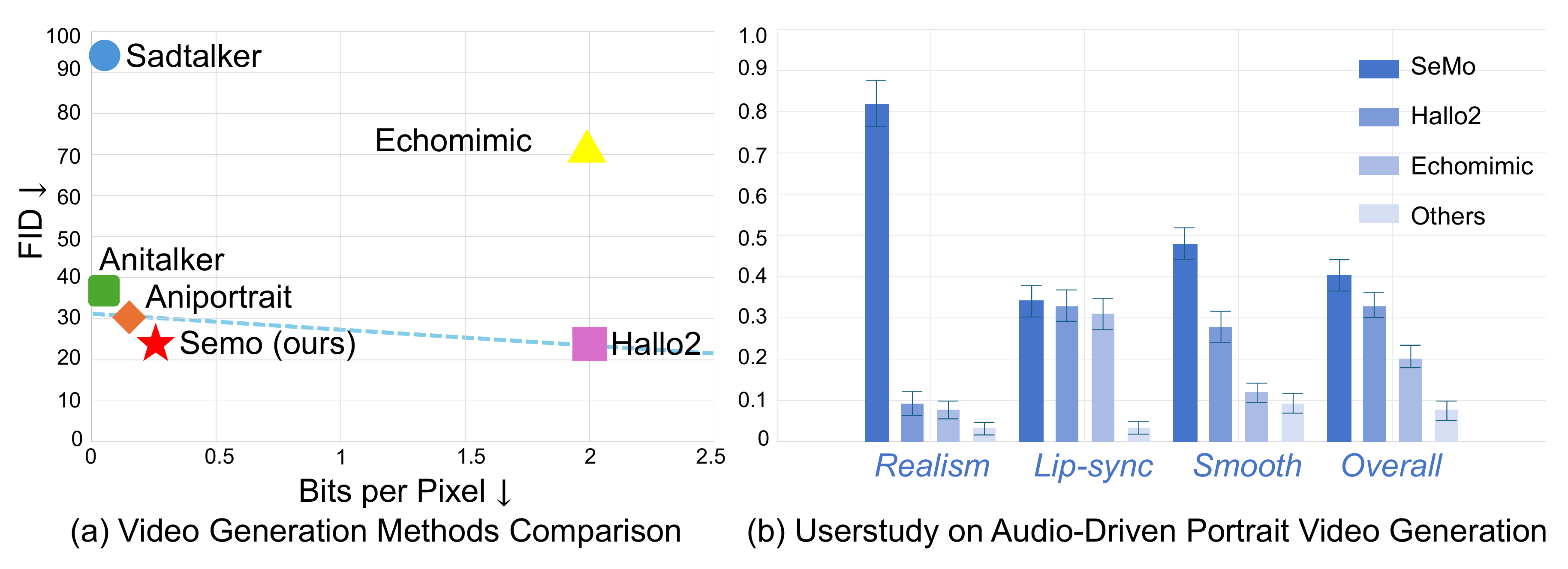}
\vspace{-6mm}
\caption{(a) Comparison of efficiency and quality among audio-driven video generation methods. Bits per Pixel is calculated as $\frac{S\cdot D}{256^2}$, where $S$ is the representation size and D is the bit width of each element, as proposed by ~\cite{sargent2025flowmodemodeseekingdiffusion}. (b) User study on audio-driven video generation. The detail procedure of user study can be found in~\cref{suppsec:userstudy}.}
\label{fig:userstudy}
\vspace{-2mm}
\end{figure}
\vspace{-2mm}

\begin{table}[t]

\caption{Video generation comparison on DH-FaceVid-1K and HDTF.}
\small\centering
\begin{minipage}[t]{0.40\textwidth}
\centering\small

\vspace{-2mm}

\resizebox{1.0\textwidth}{!}{
\begin{tabular}{lccccc}
\toprule
\multirow{2}{*}{\textbf{Method}} & \multicolumn{5}{c}{\textbf{DH-FaceVid-1K}}                                                \\ \cline{2-6} 
& \textbf{FID$\downarrow$} & \textbf{FVD$\downarrow$} & \textbf{LPIPS$\downarrow$} & \textbf{PSNR$\uparrow$} & \textbf{SSIM$\uparrow$} \\ \hline
AniPortrait~\cite{aniportrait2024}    & 44.807       & 484.861          & 0.168          & 19.103          & 0.671           \\
Hallo2~\cite{cui2024hallo2}           & 61.803       & 698.668          & 0.155          & 20.016          & 0.658   \\
SadTalker~\cite{sadtalker2023}        & 85.447       & 610.627          & 0.328          & 14.382          & 0.538   \\
Anitalker~\cite{anitalker2024}        & 56.322       & 556.607          & 0.289          & 15.064          & 0.572    \\
Echomimic~\cite{echomimic2024}        & 69.296       & 691.534          & 0.327          & 14.218          & 0.527      \\
SeMo (Ours)                           & \textbf{27.667}       & \textbf{206.523}         & \textbf{0.095}          & \textbf{22.954}          & \textbf{0.768}     \\ \bottomrule
\end{tabular}
}
\end{minipage}
~
\begin{minipage}[t]{0.55\textwidth}
\centering\small
\label{tab:gen_hdtf}
\vspace{-2mm}
\resizebox{1.0\textwidth}{!}{
\begin{tabular}{lccccccc}
\toprule
\multirow{2}{*}{\textbf{Method}} & \multicolumn{7}{c}{\textbf{HDTF}}                                                \\ \cline{2-8} 
& \textbf{FID$\downarrow$} & \textbf{FVD$\downarrow$} & \textbf{LPIPS$\downarrow$} & \textbf{PSNR$\uparrow$} & \textbf{SSIM$\uparrow$} & \textbf{Sync-C$\uparrow$} & \textbf{Sync-D$\downarrow$} \\ \hline
AniPortrait~\cite{aniportrait2024}    & 33.139       & 422.644          & 0.138          & 20.042          & 0.672          & 5.817       & 10.881     \\
Hallo2~\cite{cui2024hallo2}           & 27.447       & 276.564          & \textbf{0.105}  & 21.692          & 0.713          & \textbf{6.284}       & \textbf{8.499}     \\
SadTalker~\cite{sadtalker2023}        & 93.744       & 548.635          & 0.225          & 14.493          & 0.437          & 5.439       & 9.940     \\
Anitalker~\cite{anitalker2024}        & 41.034       & 430.082          & 0.208          & 17.434          & 0.588          & 5.927       & 8.724     \\
Echomimic~\cite{echomimic2024}        & 70.768       & 591.313          & 0.311          & 14.920          & 0.452          & 6.030       & 10.462     \\
SeMo (Ours)                           & \textbf{26.860}       & \textbf{218.414}          & 0.111          & \textbf{21.985}          & \textbf{0.739}          & 6.147       & 8.574     \\ \bottomrule
\end{tabular}
}

\end{minipage}
\label{tab:generation}
\vspace{-4mm}
\end{table}




\subsubsection{Audio-driven Video Generation.}
\label{sec:generation}

The quantitative metrics for audio-driven video generation are presented in~\cref{tab:generation}, while the qualitative (visualization) results are shown in~\cref{fig:generation}. The Sync-C and Sync-D metric was not evaluated on the DH-FaceVid-1K dataset because the models used for evaluation were not trained on Chinese datasets.
Our method not only achieves highly realistic visuals, but also excels in robustness across diverse scenarios. As shown in~\cref{fig:generation}, it effectively handles challenging cases where others often fail. Competing methods lack eyeball detail in close-ups, produce unnatural head turns, and frequently fail or generate unrealistic content in partial-face situations.
These limitations point to a common issue: prior-based methods~\cite{aniportrait2024,cui2024hallo2,facevid2024,echomimic2024} require strict assumptions about scene conditions, making such methods heavily scenario-dependent. Typically, they rely on complex data pre-processing pipelines to tightly crop and center the face within the frame. In contrast, our self-supervised approach removes the dependency on prior knowledge, enabling robust performance in unconstrained settings. As shown in~\cref{fig:userstudy}(b), our user study results demonstrate that our method achieves an impressive 81
\% win rate for perceived realism. This indicates that our approach captures more nuanced facial motions, while others often produce noticeable rigid-body translations—a subtle difference easily perceived by human evaluators but not always captured by standard quantitative metrics.In~\cref{fig:userstudy}(a), we compare the encoding efficiency of each method, which is calculated as the size of the latent representation per frame in the latent space divided by the number of pixels per frame. More visual results can be found in~\cref{suppsec:generation} and video materials.

\subsection{Ablation Study}
\label{sec:ablation}

\begin{wraptable}{r}{0.44\textwidth}
\vspace{-12mm}
  \centering
  \caption{Ablation studies on the number of motion tokens and the token channel dimension, where T denotes the number of tokens and D denotes the channel dimension.}
\resizebox{0.4\textwidth}{!}{
\begin{tabular}{lccccc}
\toprule
\multirow{2}{*}{\textbf{Token}} & \multicolumn{5}{c}{\textbf{HDTF}}                                                \\ \cline{2-6} 
& \textbf{FID$\downarrow$} & \textbf{FVD$\downarrow$} & \textbf{LPIPS$\downarrow$} & \textbf{PSNR$\uparrow$} & \textbf{SSIM$\uparrow$} \\ \hline
T1D32    & 29.877      & 237.652          & 0.068          & 26.916          & 0.824           \\
T1D64           & 26.174       & 214.236          & 0.066          & 27.520          & 0.835   \\
T1D128        & 24.467       & 192.702          & 0.053          & 27.587          & 0.837   \\
T1D256      & 22.629       & 157.235          & 0.048          & 28.430          & 0.865    \\
T1D512        & 21.431       & 136.738          & 0.043          & 29.12          & 0.871      \\
T1D768        & 21.424       & 135.678          & 0.042          & 29.385          & 0.872      \\
T1D1024        & \textbf{21.369}       & \textbf{133.547}         & \textbf{0.039}          & \textbf{29.430}          & \textbf{0.877}\\

\hdashline

T2D256      & 21.082       & 133.529          & 0.041          & 29.561          & 0.873      \\
T4D128      & 20.561       & 130.061          & 0.040          & 29.992          & 0.880      \\
T8D64       & \textbf{20.138}   & \textbf{128.390}    & \textbf{0.038}    & \textbf{30.180}      & \textbf{0.883} \\ \bottomrule
\end{tabular}
}
\label{tab:token}
\end{wraptable}
~
\noindent \textbf{Dimension and Token Number.} In \cref{tab:token}, we compared the impact of latent motion token dimensions and token numbers on model performance. For the token dimension, we fixed the number at 1 and varied its size, finding that even with a dimension as small as 32, the model could still generate reasonable images. However, smaller dimensions resulted in reduced detailed information, requiring the motion decoder to compensate, which introduced some randomness in the output, though the layout remained consistent. For the token number, we fixed the motion representation space at 512 floats and found that increasing the number of tokens improved performance. This is due to the increased parameters used by the model to analyze motion, enabling more accurate mapping from semantic tokens to low-level pixels.





\begin{figure}[t]\centering
\vspace{-2mm}
\includegraphics[width=1.0\linewidth]{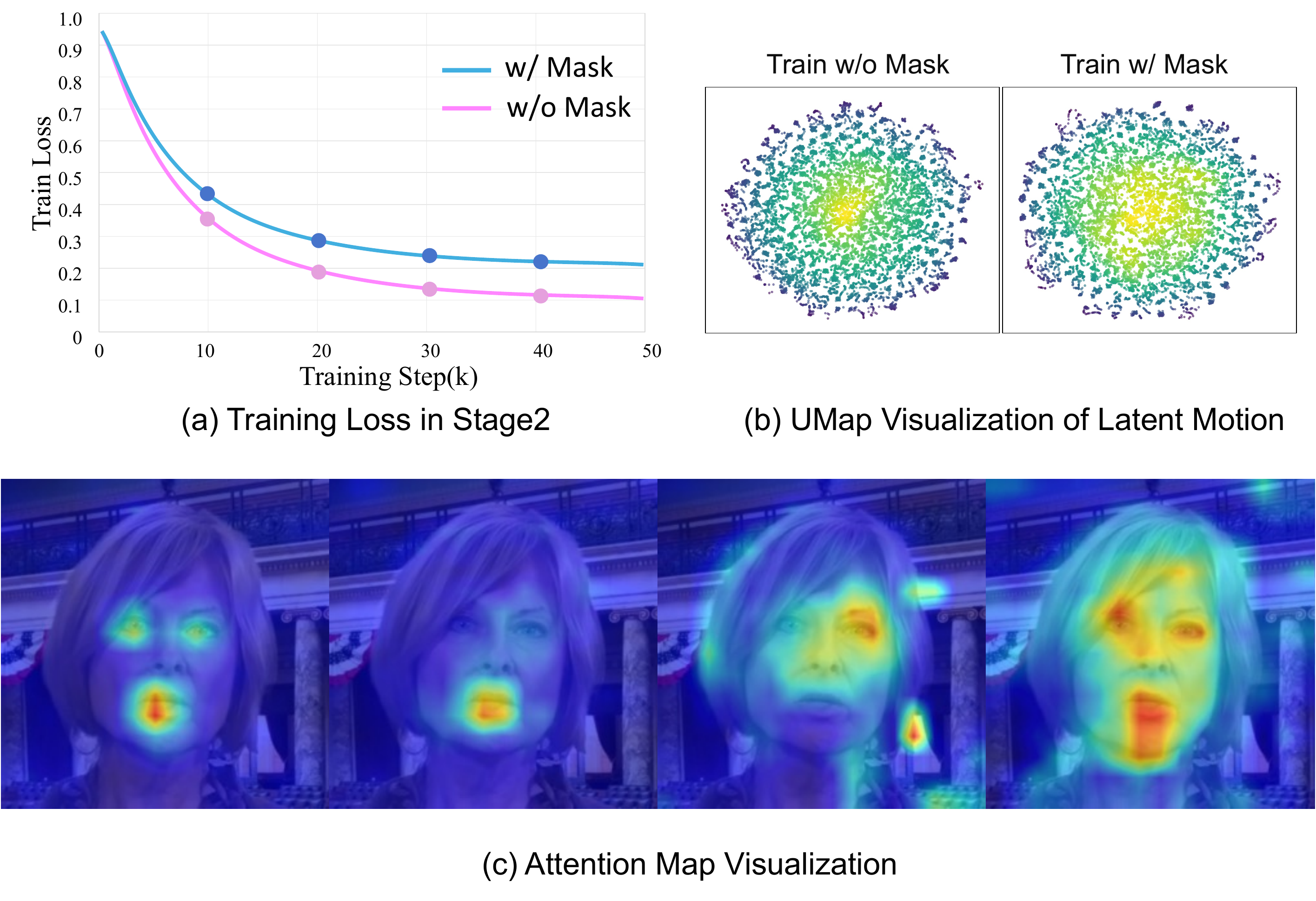}
\vspace{-4mm}
\caption{(a) Training loss of Stage 2 with and without mask.
(b) UMap~\cite{mcinnes2020umapuniformmanifoldapproximation} visualization of latent motion.
(c) Attention Map between latent motion and reference image.}
\vspace{-4mm}
\label{fig:mask}
\end{figure}

\noindent \textbf{Masking.}
\label{sec:ablation_mask}
In ~\cref{fig:mask}(a), we find that masked training accelerates Stage 2 convergence, lowers loss, and improves output quality. We infer that masking forces the model to prioritize high-level motion semantics over local details, making generation more predictable. In ~\cref{fig:mask}(b), we observe that masking encourages discrete latent motion features, aligning with findings in \cite{yao2025vavae} where discrete distributions enhance generation. In ~\cref{fig:mask}(d), We observe that, guided by the latent motion, the model mainly focuses on regions such as the eyes and mouth, where the movements are more significant. In ~\cref{suppfig:mask}(c), we observe that using a masking strategy can partially separate identity and motion: replacing the reference video with another identity shows less identity leakage when masking is applied. We attribute this effect to the randomly sampled mask ratios, which encourage the model to abstract at high ratios and retain some facial details at lower ones.

%% file: sections/6_conclusion.tex
\section{Conclusion and Limitations}
\label{sec:conclusion}

\vspace{-2mm}

Portrait video generation has a wide range of applications, but it faces challenges in both generation efficiency and quality. Moreover, existing methods often rely on prior knowledge, which limits their applicability to fixed facial scenarios. In this work, we address these issues by proposing a self-supervised motion representation framework \textbf{SeMo}, which is clear and straightforward, consisting of three steps: Abstraction, Reasoning, and Generation. Thanks to the compact and expressive nature of SeMo, we demonstrate that SeMo significantly improves generation efficiency and quality. Beyond portrait video generation, our pipeline is broadly applicable to general video generation tasks. In future work, we plan to further explore its feasibility and effectiveness across a wider range of scenarios.

\noindent \textbf{Limitation.}
Due to the need for efficient generation and better user accessibility, we limited our experiments to a resolution of 256×256. However, this downsampling, together with compression by the SD VAE, causes some information loss, resulting in blurry teeth and flickering in high-frequency regions such as beards. In future work, we plan to explore higher resolutions to further improve the visual quality of the generated results.

%% file: sections/7_supplement.tex
\newpage

\appendix

\section{Motivation}
\label{suppsec:motivation}
Our motivation stems from the observation that video reasoning is inherently high-dimensional and complex; predicting every pixel is highly uncertain, and humans do not explicitly forecast each pixel in their brains. Instead, we tend to abstract concepts from what we see and perform efficient reasoning in this conceptual space, which not only improves the reliability but also the efficiency of predictions. Inspired by this, we apply a similar principle to portrait video generation. Rather than directly predicting pixels, the model first abstracts each frame into high-level concepts, enabling more effective prediction within this compact space. Leveraging the strong generative and conditional capabilities of diffusion models, we then progressively refine predictions from high-level semantic concepts to final pixel-level outputs in a coarse-to-fine manner.

\begin{figure}[H]
    \centering
    \includegraphics[width=1\textwidth]{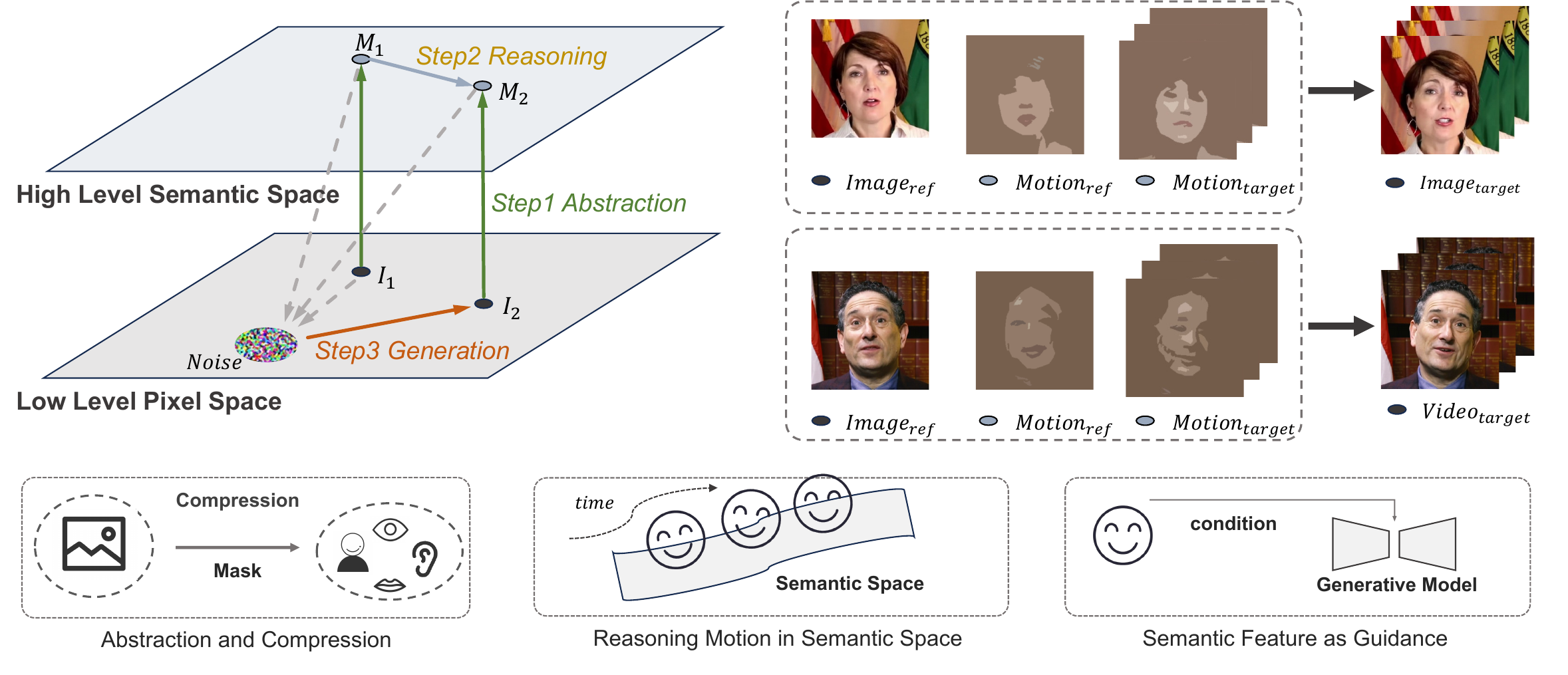}
    \caption{\textbf{Motivation.}}
    \vspace{-2mm}
    \label{suppfig:motivation}
\end{figure}

As illustrated in the~\ref{suppfig:motivation}, given a reference image, reference motion, and target motion, the model can easily generate the target frame at the pixel level. To this end, we combine self-supervised learning and diffusion-based generative modeling in a three-step framework: \textbf{Abstraction}, \textbf{reasoning}, and \textbf{generation}.

\section{Implementation Details}
\label{suppsec:detail}
As described in Sec.~\ref{sec:encode}, SeMo adopts an EncoderDecoder architecture. Specifically, In the motion encoder, we employ 8 Transformer layers. The motion decoder consists of 12 denoising layers, including both Motion Control Layers and Temporal Align Layers. For the motion generator, we use 8 DiT layers. Throughout all modules, the hidden dimension is set to 768 and we use 8 attention heads. The Motion Autoencoder comprises 450 million parameters, and the motion generator comprises 200 million parameters. For input preprocessing, all images and video frames are resized to $256 \times 256$ resolution and pixel values are normalized to the [-1,1] range. All models are trained under similar settings: a base learning rate of $10^{-4}$ per 64 batch size and the AdamW optimizer with $\beta_1=0.9$, $\beta_2=0.95$, $\text{decay}=0.05$. Learning rate is set to $1e{-4}$. In Stage1, Models are trained with a batch size of 8 on 8 NVIDIA A800 GPUs for 300K iterations. In Stage2, Models are trained with a batch size of 8 on 8 NVIDIA A800 GPUs for 200K iterations. We employ gradient clipping with a threshold of 1.0 to stabilize training.

\begin{table}[h!]
\centering
\caption{Training Configuration}
\label{tab:training_config}
\begin{tabular}{l|l}
\toprule[1.5pt]
\textbf{Parameter}              & \textbf{Value} \\ \midrule
Input Video Resolution          & 256            \\ \midrule
Input Video Frames              & 16             \\ \midrule
Input Video FPS                 & 25              \\ \midrule
Optimizer                       & Adam; $\beta_1 = 0.9, \beta_2 = 0.95$ \\ \midrule
Learning Rate                   & $1 \times 10^{-4}$ \\ \midrule
Warmup Steps                    & 5000           \\ \midrule
Learning Rate Scheduler         & Cosine Annealing \\ \midrule
$\mathcal{L}_{KL}$ & 0.001  \\ \midrule
Weight Decay & 0.0001 \\ \midrule
Training Batch Size             & 8              \\ \midrule
Training Device                 & 8 $\times$ 80G A800 GPUs \\ 
\bottomrule[1.5pt]
\end{tabular}
\end{table}

\section{Userstudy}
\label{suppsec:userstudy}
We randomly selected 10 images from HDTF and DH-FaceVid-1K, and generated 20 videos using 2 random audio clips. We invited 50 participants, divided the participants into 5 groups, each participant was asked to select the best model from 4 different perspectives (realism, lip-sync, smooth, overall).  We significantly outperform competitors in realism and match the state-of-the-art (SOTA) in lip sync, smoothness, and overall quality.
\section{Additional Analysis}
\label{suppsec:Analysis}

\subsection{Analysis on Mask }
\label{suppsec:mask}

\begin{figure}[t]
    \centering
    \includegraphics[width=1\textwidth]{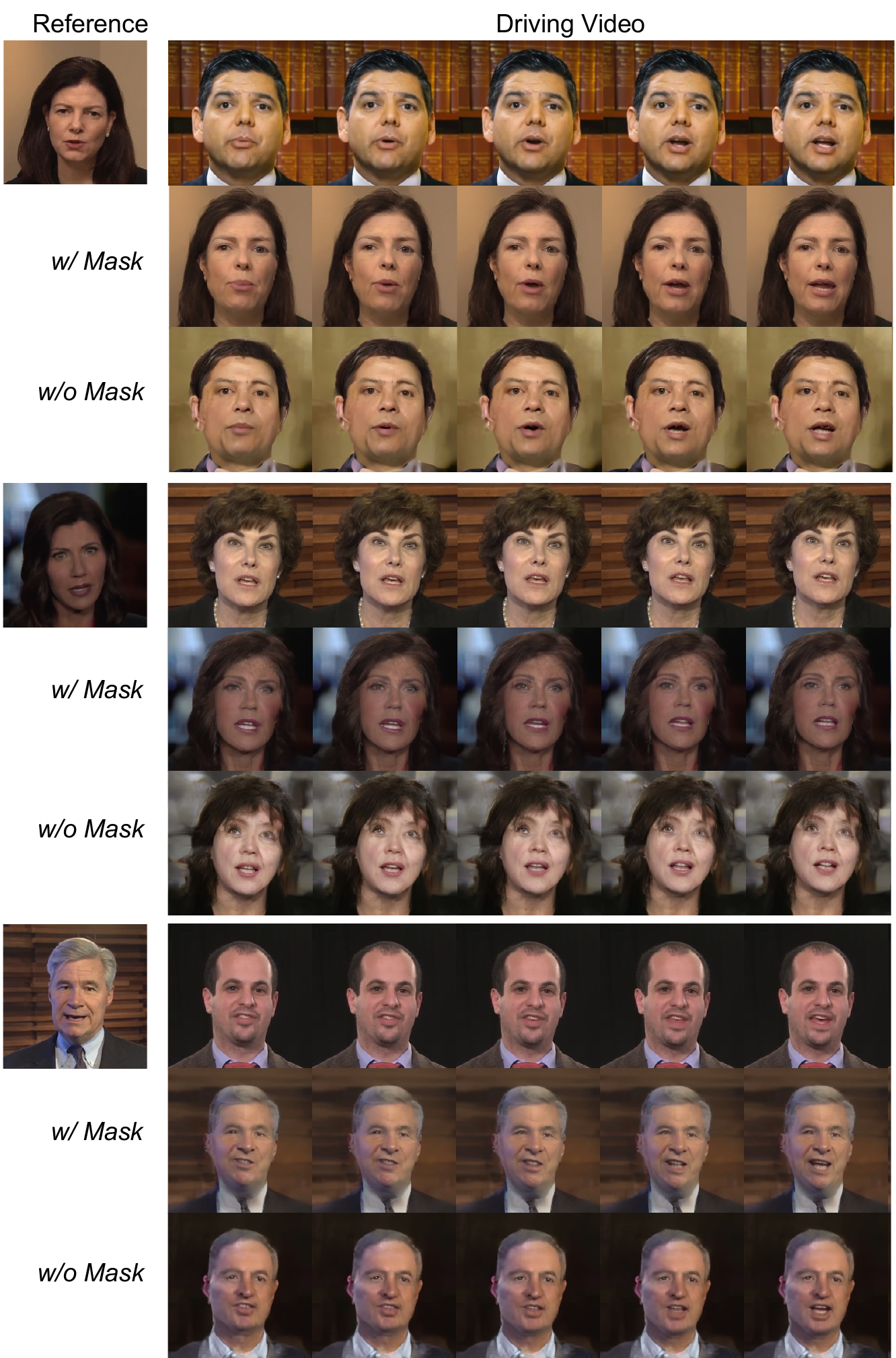}
    \caption{Cross Identity Driving Example}
    \vspace{-2mm}
    \label{suppfig:mask}
\end{figure}

Additional results for the cross-ID driving task are shown in~\cref{suppfig:mask}, demonstrating that the mask strategy significantly suppresses identity leakage, although complete disentanglement is still not achieved. However, we emphasize that such disentanglement is not strictly necessary for video generation tasks. Excessive retention of fine-grained details may hinder the generation process in the second stage, while discarding unnecessary details helps stabilize both training and inference.

\begin{figure}[t]
    \centering
    \includegraphics[width=1\textwidth]{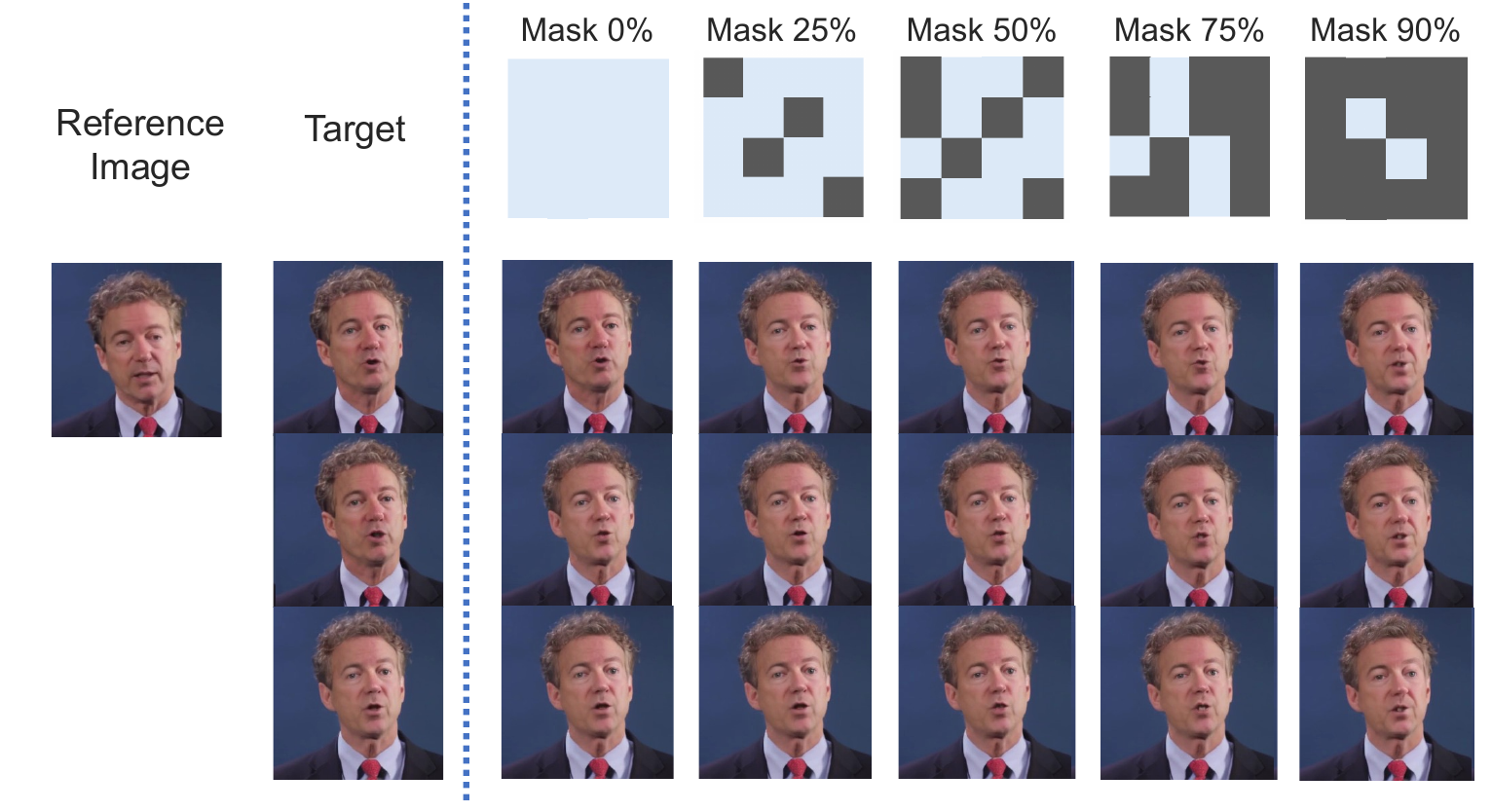}
    \caption{Reconstruction with Different Mask Ratio.}
    \vspace{-2mm}
    \label{suppfig:mask2}
\end{figure}

As shown in ~\cref{suppfig:mask2}, even when 90\% of the content is masked, the model is still able to reconstruct reasonable frames. This indicates a high degree of redundancy in motion information, suggesting that effective motion modeling can be achieved with only a small fraction of inputs.

\begin{figure}[t]
    \centering
    \includegraphics[width=1\textwidth]{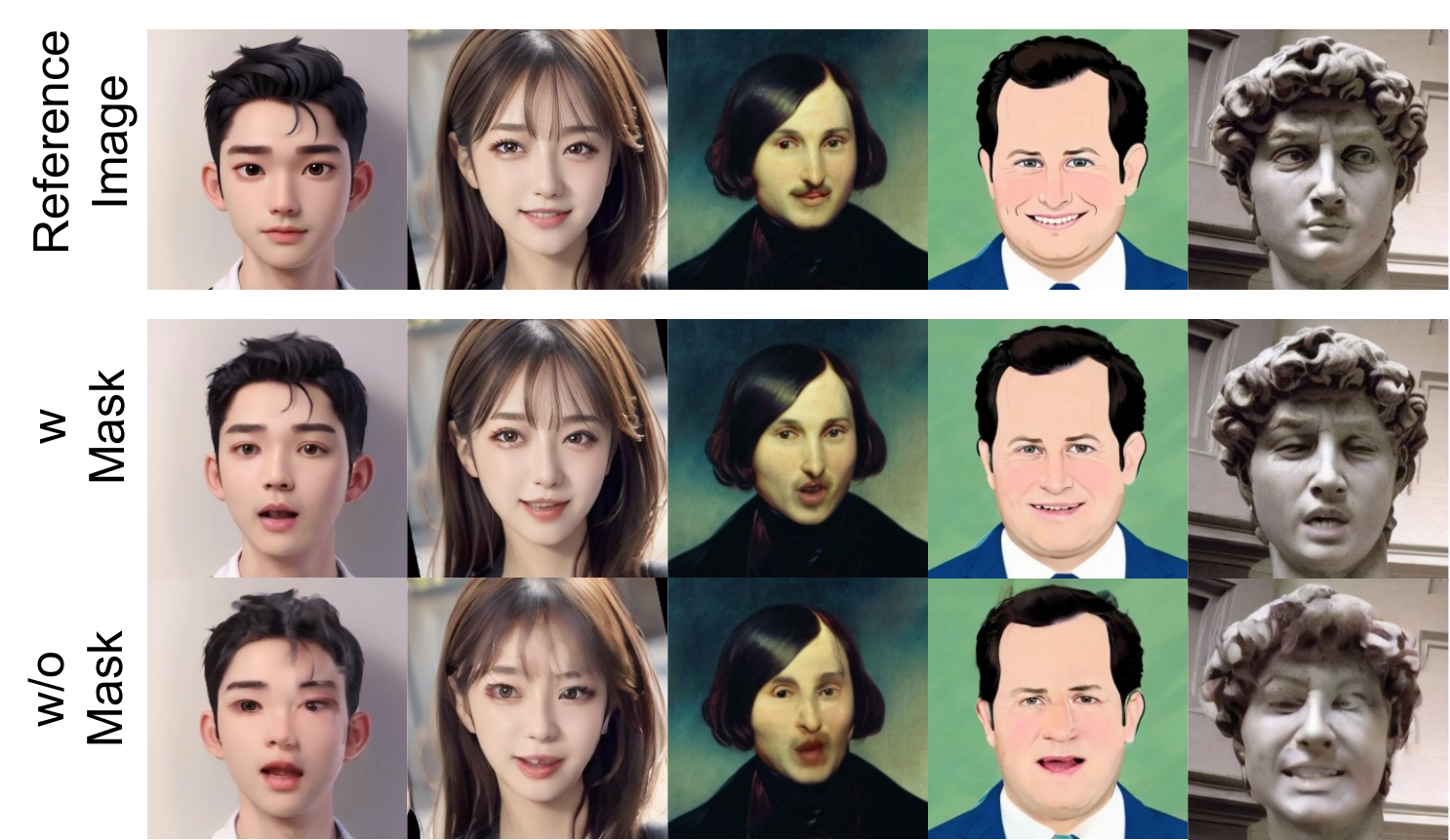}
    \caption{Result of the analysis on generalization.The mask strategy enhances the model’s generalization ability.}
    \vspace{-2mm}
    \label{suppfig:mask3}
\end{figure}

Furthermore, ~\cref{suppfig:mask3} reveals that without the masking strategy, the model tends to memorize excessive details, which compromises its ability to generalize to out-of-distribution data. This observation further highlights the importance of abstract representation for robust video generation.

\subsection{More Visualization Result of Attention Map}
\label{suppsec:atten}

\begin{figure}[t]
    \centering
    \includegraphics[width=1\textwidth]{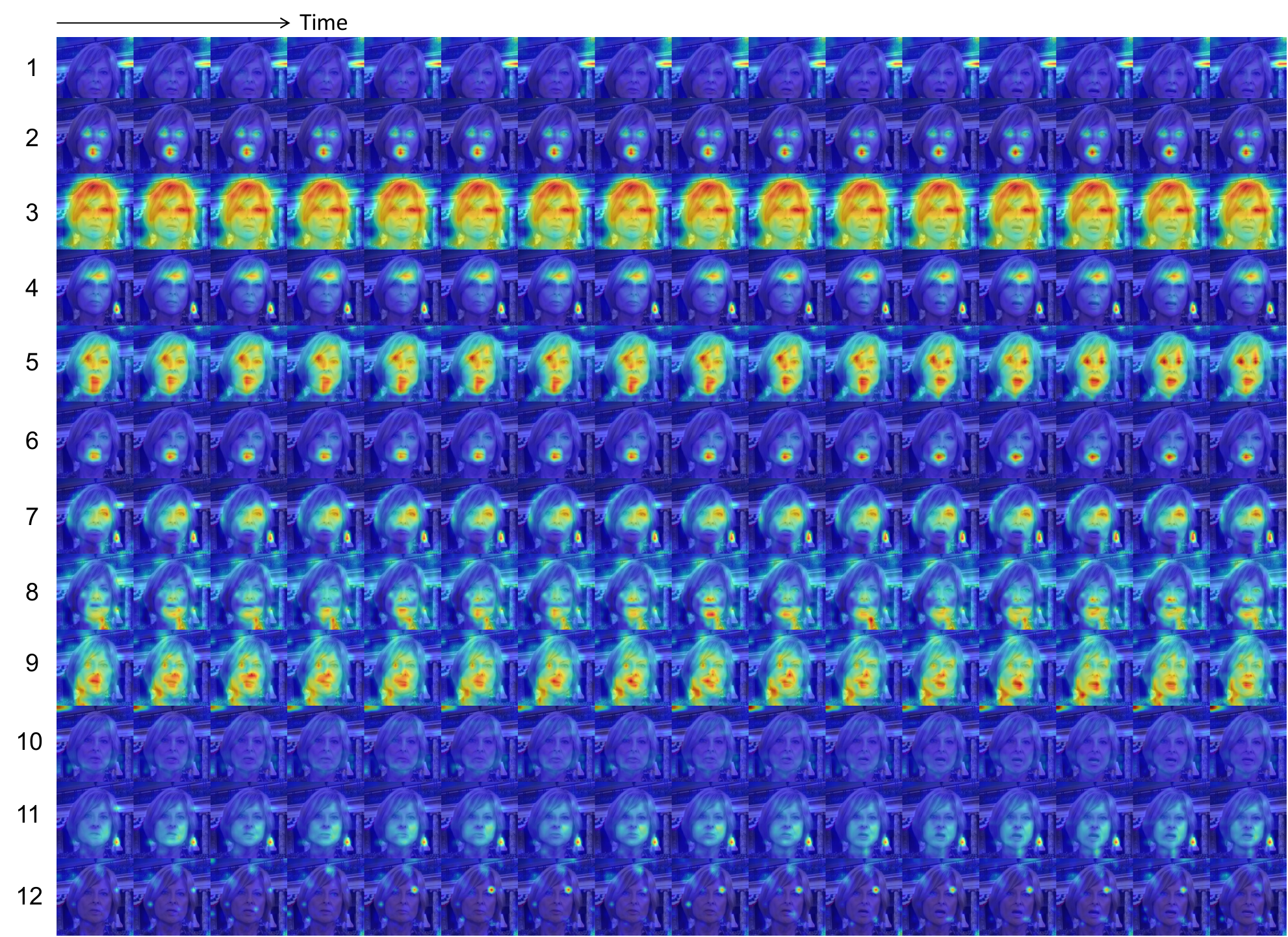}
    \caption{Visualization of Attention Map.}
    \vspace{-2mm}
    \label{suppfig:atten}
\end{figure}

In~\cref{suppfig:atten}, We visualized the attention map between the motion latent and the reference image in the decoder. The results show that the interaction between the motion latent and the reference image mainly focuses on facial regions, especially those areas that typically undergo significant movement. In addition to the face, the model also attends to local regions prone to motion, such as earrings. By contrast, the model pays less attention to the background and other regions that are unlikely to move. This phenomenon indicates that the motion latent contains rich and detailed motion information, and the model can effectively capture and utilize these highly motion-relevant local features, thereby enabling precise motion generation in target areas.

\subsection{More Visualization Result of Audio-driven Video Generation}
\label{suppsec:generation}

\begin{figure}[t]
    \centering
    \includegraphics[width=1\textwidth]{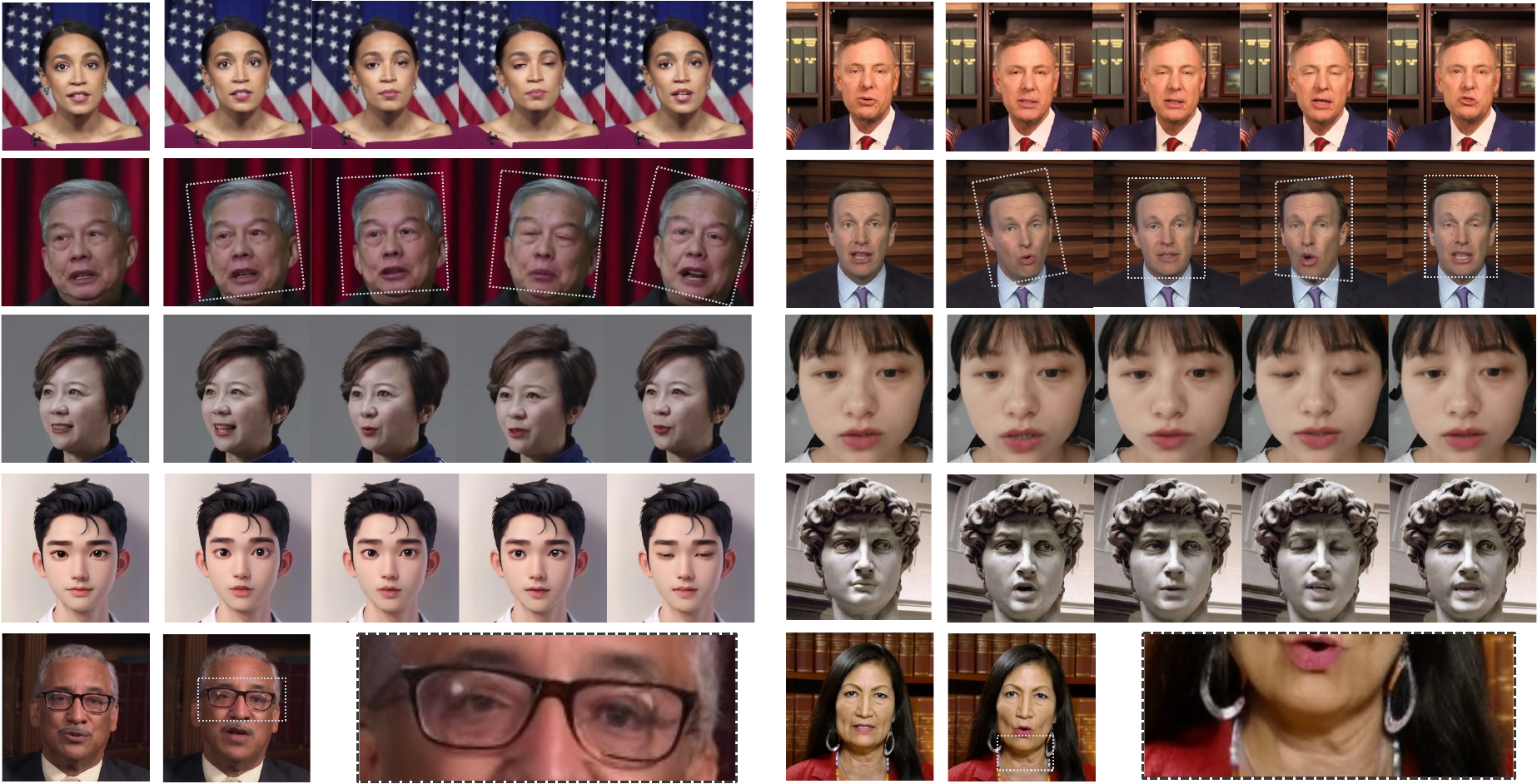}
    \caption{Visualization of Audio-driven Portrait Video Generation.}
    \vspace{-2mm}
    \label{suppfig:talking1}
\end{figure}

\begin{figure}[t]
    \centering
    \includegraphics[width=1\textwidth]{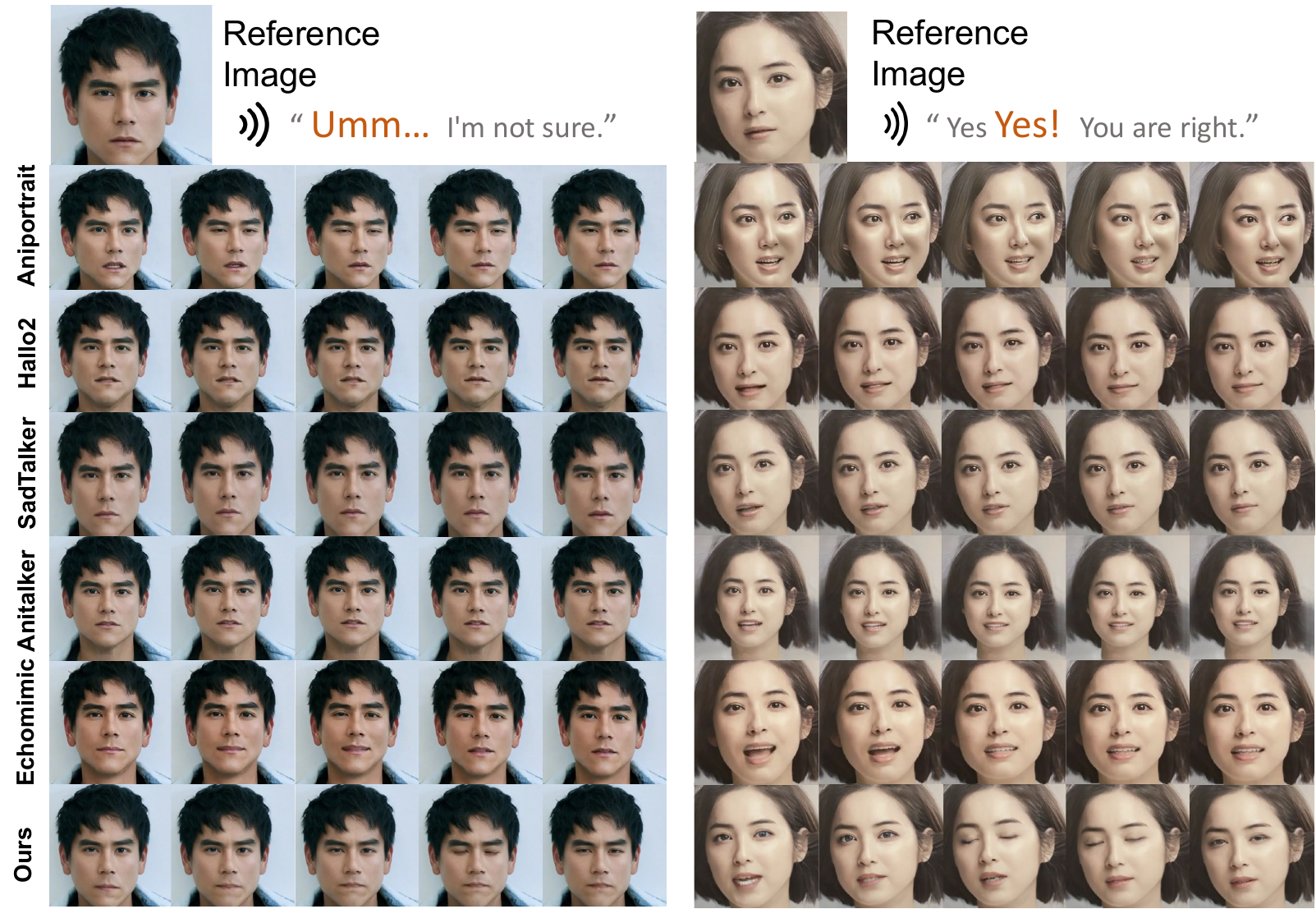}
    \caption{Visualization of Audio-driven Portrait Video Generation. It can also produce natural movements—such as blinking—even during interjections or periods of silence}
    \vspace{-2mm}
    \label{suppfig:talking2}
\end{figure}

We provide more experimental results of talking head in~\cref{suppfig:talking1}~\cref{suppfig:talking2}. In ~\cref{suppfig:talking2}, we observe that in extreme speech cases, such as silence, the labial sound 'emmm' or certain interjections, other methods remain static, whereas our method exhibits natural behaviors, such as blinking during silence and producing natural mouth movements during interjections.

\subsection{Transfer to general video reconstruction tasks}
\label{suppsec:general}
\begin{figure}[t]
    \centering
    \includegraphics[width=1\textwidth]{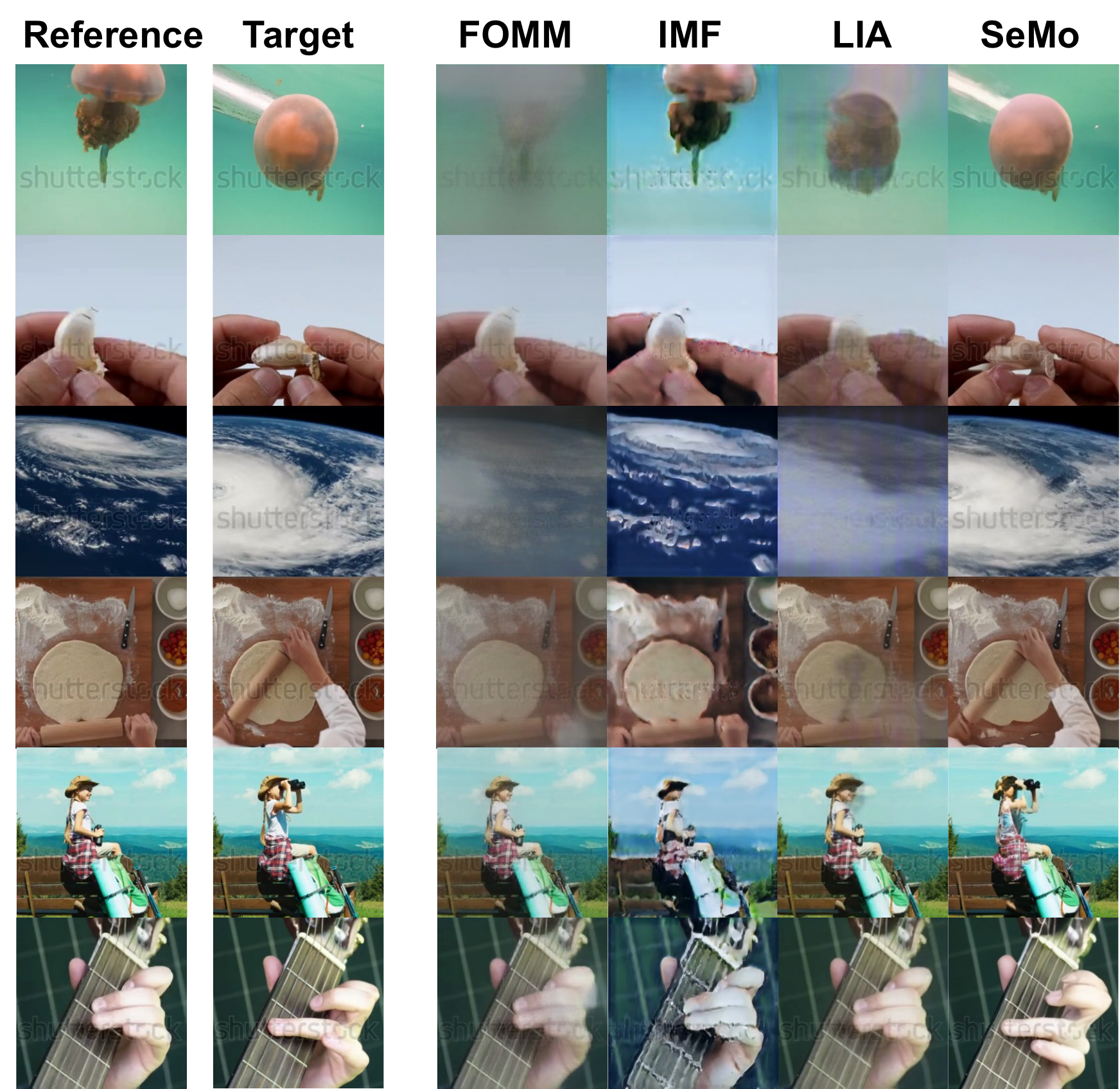}
    \caption{Transfer to general video reconstruction tasks.}
    \vspace{-2mm}
    \label{suppfig:general}
\end{figure}

Since our framework is fully self-supervised, in~\cref{suppfig:general}, we further explored its applicability to general video scenarios. We found that our method is suitable for general video scenes with clear subjects and relatively simple motion. However, in scenarios with multiple or complex subjects, trailing artifacts may occur, as shown in the figure. Compared to similar methods, our approach achieves better performance on general video datasets. In our experiments, we reproduced the IMF method and trained it on the WebVid dataset.

\section{Novelty}
\label{suppsec:novelty}
The work most closely related to ours is IMF~\cite{imf2024}. However, our objectives and implementation strategies are fundamentally different.
First, regarding objectives, IMF focuses solely on image-to-image transfer, whereas our goal is to transfer from images to videos. Moreover, the way we abstract and represent motion also differs: we employ a transformer architecture along with a masking strategy, enabling the model to encode compact and semantically rich motion features.
Second, our decoding approach is distinct. We use rectified flow model as the decoder, which leads to a simpler loss design and achieves better visual results.
Finally, we demonstrate the effectiveness of our method on the challenging downstream task of audio-driven video generation.



\newpage

%% file: sections/8_checklist.tex
\newpage

\section*{NeurIPS Paper Checklist}

\begin{enumerate}

\item {\bf Claims}
    \item[] Question: Do the main claims made in the abstract and introduction accurately reflect the paper's contributions and scope?
    \item[] Answer: \answerYes{} 
    \item[] Justification:  This paper proposes a self-supervised learning based framework for portrait motion representation, which improves the efficiency and quality of portrait video generation. The proposed algorithm are illustrated in Section~\ref{sec:method}. Experimental results are illustrated in Section~\ref{sec:exp}.

\item {\bf Limitations}
    \item[] Question: Does the paper discuss the limitations of the work performed by the authors?
    \item[] Answer: \answerYes{} 
    \item[] Justification: We include the limitations of our work in Section~\ref{sec:conclusion}.

\item {\bf Theory assumptions and proofs}
    \item[] Question: For each theoretical result, does the paper provide the full set of assumptions and a complete (and correct) proof?
    \item[] Answer: \answerNA{} 
    \item[] Justification: This is not a theoretical paper.

    \item {\bf Experimental result reproducibility}
    \item[] Question: Does the paper fully disclose all the information needed to reproduce the main experimental results of the paper to the extent that it affects the main claims and/or conclusions of the paper (regardless of whether the code and data are provided or not)?
    \item[] Answer: \answerYes{} 
    \item[] Justification: Datasets, models, and hyperparameters used in implementing proposed algorithms are all described in detail. See Section~\ref{sec:exp} and Appendix.

\item {\bf Open access to data and code}
    \item[] Question: Does the paper provide open access to the data and code, with sufficient instructions to faithfully reproduce the main experimental results, as described in supplemental material?
    \item[] Answer: \answerYes{} 
    \item[] Justification: We will provide the code to train and evaluate the proposed algorithm, which reproduces the experiment results in the paper after the rebuttal process.

\item {\bf Experimental setting/details}
    \item[] Question: Does the paper specify all the training and test details (e.g., data splits, hyperparameters, how they were chosen, type of optimizer, etc.) necessary to understand the results?
    \item[] Answer: \answerYes{} 
    \item[] Justification: We provide the details of our experiments in Section~\ref{sec:exp} and Appendix.

\item {\bf Experiment statistical significance}
    \item[] Question: Does the paper report error bars suitably and correctly defined or other appropriate information about the statistical significance of the experiments?
    \item[] Answer: \answerYes{} 
    \item[] Justification: We report standard error.

\item {\bf Experiments compute resources}
    \item[] Question: For each experiment, does the paper provide sufficient information on the computer resources (type of compute workers, memory, time of execution) needed to reproduce the experiments?
    \item[] Answer: \answerYes{} 
    \item[] Justification: We report the compute resources during training in Section~\ref{sec:exp} and Appendix.
    
\item {\bf Code of ethics}
    \item[] Question: Does the research conducted in the paper conform, in every respect, with the NeurIPS Code of Ethics \url{https://neurips.cc/public/EthicsGuidelines}?
    \item[] Answer: \answerYes{} 
    \item[] Justification: We followed the NeurIPS Code of Ethics.

\item {\bf Broader impacts}
    \item[] Question: Does the paper discuss both potential positive societal impacts and negative societal impacts of the work performed?
    \item[] Answer: \answerNA{} 
    \item[] Justification: This work focuses on a academic, publicly-available benchmark. This work is not related to any private or personal data, and there’s no explicit negative social impacts.
    
\item {\bf Safeguards}
    \item[] Question: Does the paper describe safeguards that have been put in place for responsible release of data or models that have a high risk for misuse (e.g., pretrained language models, image generators, or scraped datasets)?
    \item[] Answer: \answerNA{} 
    \item[] Justification: our work poses no risk of misuse.

\item {\bf Licenses for existing assets}
    \item[] Question: Are the creators or original owners of assets (e.g., code, data, models), used in the paper, properly credited and are the license and terms of use explicitly mentioned and properly respected?
    \item[] Answer: \answerYes{} 
    \item[] Justification: This paper cite the original papers such as dataset.

\item {\bf New assets}
    \item[] Question: Are new assets introduced in the paper well documented and is the documentation provided alongside the assets?
    \item[] Answer: \answerYes{} 
    \item[] Justification: This paper will release code for running experiments and it is well documented.

\item {\bf Crowdsourcing and research with human subjects}
    \item[] Question: For crowdsourcing experiments and research with human subjects, does the paper include the full text of instructions given to participants and screenshots, if applicable, as well as details about compensation (if any)? 
    \item[] Answer: \answerNA{} 
    \item[] Justification: This work does not involve crowdsourcing nor research with human subjects.


\item {\bf Institutional review board (IRB) approvals or equivalent for research with human subjects}
    \item[] Question: Does the paper describe potential risks incurred by study participants, whether such risks were disclosed to the subjects, and whether Institutional Review Board (IRB) approvals (or an equivalent approval/review based on the requirements of your country or institution) were obtained?
    \item[] Answer: \answerNA{} 
    \item[] Justification: The paper does not involve crowdsourcing nor research with human subjects.

\item {\bf Declaration of LLM usage}
    \item[] Question: Does the paper describe the usage of LLMs if it is an important, original, or non-standard component of the core methods in this research? Note that if the LLM is used only for writing, editing, or formatting purposes and does not impact the core methodology, scientific rigorousness, or originality of the research, declaration is not required.
    \item[] Answer: \answerNA{} 
    \item[] Justification: The LLM is used only for writing.

\end{enumerate}

%% file: main.bbl
\begin{thebibliography}{10}

\bibitem{HumanVideo2018}
Haoye Cai and Chunyan Bai.
\newblock Deep video generation, prediction and completion of human action sequences.
\newblock In {\em ECCV}, 2018.

\bibitem{everydance2019}
Caroline Chan and Shiry Ginosar.
\newblock Everybody dance now.
\newblock In {\em ICCV}, 2019.

\bibitem{moco2021}
Xinlei Chen, Saining Xie, and Kaiming He.
\newblock An empirical study of training self-supervised vision transformers.
\newblock In {\em ICCV}, October 2021.

\bibitem{echomimic2024}
Zhiyuan Chen and Jiajiong Cao.
\newblock Echomimic: Lifelike audio-driven portrait animations through editable landmark conditioning.
\newblock In {\em AAAI}, 2024.

\bibitem{cho2024sora}
Joseph Cho and Fachrina~Dewi Puspitasari.
\newblock Sora as an agi world model? a complete survey on text-to-video generation.
\newblock {\em arXiv}, 2024.

\bibitem{cui2024hallo2}
Jiahao Cui and Hui Li.
\newblock Hallo2: Long-duration and high-resolution audio-driven portrait image animation.
\newblock In {\em ICLR}, 2025.

\bibitem{facevid2024}
Donglin Di and He~Feng.
\newblock Facevid-1k: A large-scale high-quality multiracial human face video dataset, 2024.

\bibitem{vit2021}
Alexey Dosovitskiy, Lucas Beyer, and Alexander Kolesnikov.
\newblock An image is worth 16x16 words: Transformers for image recognition at scale.
\newblock In {\em ICLR}, 2021.

\bibitem{imf2024}
Yue Gao, Jiahao Li, and Lei Chu.
\newblock Implicit motion function.
\newblock In {\em CVPR}, 2024.

\bibitem{Girdhar2023OmniMAE}
Rohit Girdhar and Alaaeldin El-Nouby.
\newblock Omnimae: Single model masked pretraining on images and videos.
\newblock In {\em CVPR}, 2023.

\bibitem{guo2024liveportrait}
Jianzhu Guo and Dingyun Zhang.
\newblock Liveportrait: Efficient portrait animation with stitching and retargeting control.
\newblock {\em arXiv preprint arXiv:2407.03168}, 2024.

\bibitem{guo2023animatediff}
Yuwei Guo and Ceyuan Yang.
\newblock Animatediff: Animate your personalized text-to-image diffusion models without specific tuning.
\newblock {\em ICLR}, 2024.

\bibitem{mae2022}
Kaiming He and Xinlei Chen.
\newblock Masked autoencoders are scalable vision learners.
\newblock In {\em CVPR}, Jun 2022.

\bibitem{gaia2024}
Tianyu He and Junliang Guo.
\newblock Gaia: Zero-shot talking avatar generation.
\newblock In {\em ICLR}, 2024.

\bibitem{heusel2017gans}
Martin Heusel and Ramsauer.
\newblock Gans trained by a two time-scale update rule converge to a local nash equilibrium.
\newblock In {\em NeurIPS}, volume~30, 2017.

\bibitem{li2024survey}
Chengxuan Li and Di~Huang.
\newblock A survey on long video generation: Challenges, methods, and prospects.
\newblock {\em arXiv}, 2024.

\bibitem{rcg2024}
Tianhong Li, Dina Katabi, and Kaiming He.
\newblock Return of unconditional generation: A self-supervised representation generation method.
\newblock In {\em NeurIPS}, 2024.

\bibitem{liu2025moee}
Huaize Liu and Wenzhang Sun.
\newblock Moee: Mixture of emotion experts for audio-driven portrait animation.
\newblock In {\em CVPR}, 2024.

\bibitem{anitalker2024}
Tao Liu and Feilong Chen.
\newblock Anitalker: animate vivid and diverse talking faces through identity-decoupled facial motion encoding.
\newblock In {\em ACM MM}, 2024.

\bibitem{MODA2023}
Yunfei Liu, Lijian Lin, and Yu~Fei.
\newblock Moda: Mapping-once audio-driven portrait animation with dual attentions.
\newblock In {\em CVPR}, 2023.

\bibitem{Latte2024}
Xin Ma and Yaohui Wang.
\newblock Latte: Latent diffusion transformer for video generation.
\newblock {\em arXiv}, 2024.

\bibitem{mcinnes2020umapuniformmanifoldapproximation}
Leland McInnes, John Healy, and James Melville.
\newblock Umap: Uniform manifold approximation and projection for dimension reduction, 2020.

\bibitem{wav2lip2020}
K~R Prajwal and Mukhopadhyay.
\newblock A lip sync expert is all you need for speech to lip generation in the wild.
\newblock In {\em ACM MM}, MM ’20, 2020.

\bibitem{qian2021spatiotemporal}
Rui Qian and Tianjian Meng.
\newblock Spatiotemporal contrastive video representation learning.
\newblock In {\em CVPR}, 2021.

\bibitem{whisper2022}
Alec Radford, Jong~Wook Kim, and Tao Xu.
\newblock Robust speech recognition via large-scale weak supervision, 2022.

\bibitem{sapiens2024}
Khirodkar Rawal and Bagautdinov.
\newblock Sapiens: Foundation for human vision models.
\newblock In {\em ECCV}, 2024.

\bibitem{sd2022}
Robin Rombach and Andreas Blattmann.
\newblock High-resolution image synthesis with latent diffusion models.
\newblock In {\em CVPR}, 2022.

\bibitem{sargent2025flowmodemodeseekingdiffusion}
Kyle Sargent, Kyle Hsu, Justin Johnson, Li~Fei-Fei, and Jiajun Wu.
\newblock Flow to the mode: Mode-seeking diffusion autoencoders for state-of-the-art image tokenization, 2025.

\bibitem{fomm2020}
Aliaksandr Siarohin and Lathuilière.
\newblock First order motion model for image animation.
\newblock In {\em NeurIPS}, December 2019.

\bibitem{mraa2021}
Aliaksandr Siarohin, Oliver Woodford, and Ren.
\newblock Motion representations for articulated animation.
\newblock In {\em CVPR}, 2021.

\bibitem{sun2024uniavatar}
Wenzhang Sun and Xiang Li.
\newblock Uniavatar: Taming lifelike audio-driven talking head generation with comprehensive motion and lighting control.
\newblock {\em arXiv preprint arXiv:2412.19860}, 2024.

\bibitem{vividtalk2023}
Xusen Sun, Longhao Zhang, and Hao Zhu.
\newblock Vividtalk: One-shot audio-driven talking head generation based 3d hybrid prior.
\newblock {\em arXiv preprint arXiv:2312.01841}, 2023.

\bibitem{emo2024}
Linrui Tian, Qi~Wang, and Bang Zhang.
\newblock Emo: Emote portrait alive - generating expressive portrait videos with audio2video diffusion model under weak conditions.
\newblock In {\em ECCV}, 2024.

\bibitem{tian2024reducio}
Rui Tian, Qi~Dai, Jianmin Bao, Kai Qiu, Yifan Yang, Chong Luo, Zuxuan Wu, and Yu-Gang Jiang.
\newblock Reducio! generating 1024$times$1024 video within 16 seconds using extremely compressed motion latents.
\newblock {\em arXiv preprint arXiv:2411.13552}, 2024.

\bibitem{videomae2022}
Zhan Tong, Yibing Song, and Jue Wang.
\newblock Video{MAE}: Masked autoencoders are data-efficient learners for self-supervised video pre-training.
\newblock In {\em NeurIPS}, 2022.

\bibitem{kaisiyuan2020mead}
Kaisiyuan Wang and Wu.
\newblock Mead: A large-scale audio-visual dataset for emotional talking-face generation.
\newblock In {\em ECCV}, 8 2020.

\bibitem{wang2023VideoMAE}
Limin Wang and Bingkun Huang.
\newblock Videomae v2: Scaling video masked autoencoders with dual masking.
\newblock In {\em CVPR}, 2023.

\bibitem{wang2018videotovideosynthesis}
Ting-Chun Wang and Ming-Yu Liu.
\newblock Video-to-video synthesis.
\newblock In {\em NeurIPS}, 2018.

\bibitem{lia2024}
Yaohui Wang, Di~Yang, and Francois Bremond.
\newblock Lia: Latent image animator.
\newblock {\em IEEE TPAMI}, 2024.

\bibitem{wang2024vidtwin}
Yuchi Wang, Junliang Guo, Xinyi Xie, Tianyu He, Xu~Sun, and Jiang Bian.
\newblock Vidtwin: Video vae with decoupled structure and dynamics.
\newblock {\em arXiv preprint arXiv:2412.17726}, 2024.

\bibitem{wang2004image}
Zhou Wang and Bovik.
\newblock Image quality assessment: from error visibility to structural similarity.
\newblock {\em IEEE TIP}, 13(4), 2004.

\bibitem{aniportrait2024}
Huawei Wei, Zejun Yang, and Zhisheng Wang.
\newblock Aniportrait: Audio-driven synthesis of photorealistic portrait animation, 2024.

\bibitem{vasa2024}
Sicheng Xu and Guojun Chen.
\newblock Vasa-1: Lifelike audio-driven talking faces generated in real time.
\newblock In {\em NeurIPS}, 2024.

\bibitem{posevideo2018}
Ceyuan Yang and Zhe Wang.
\newblock Pose guided human video generation.
\newblock In {\em ECCV}, 2018.

\bibitem{yang2020video}
Ceyuan Yang and Yinghao Xu.
\newblock Video representation learning with visual tempo consistency.
\newblock {\em arXiv preprint arXiv:2006.15489}, 2020.

\bibitem{yao2025vavae}
Jingfeng Yao, Bin Yang, and Xinggang Wang.
\newblock Reconstruction vs. generation: Taming optimization dilemma in latent diffusion models.
\newblock In {\em CVPR}, 2025.

\bibitem{tiktok2024}
Qihang Yu, Mark Weber, and Xueqing Deng.
\newblock An image is worth 32 tokens for reconstruction and generation.
\newblock In {\em ICLR}, 2024.

\bibitem{yu2025repa}
Sihyun Yu, Sangkyung Kwak, Huiwon Jang, Jongheon Jeong, Jonathan Huang, Jinwoo Shin, and Saining Xie.
\newblock Representation alignment for generation: Training diffusion transformers is easier than you think.
\newblock In {\em ICLR}, 2025.

\bibitem{yu2024efficient}
Sihyun Yu, Weili Nie, De-An Huang, Boyi Li, Jinwoo Shin, and Anima Anandkumar.
\newblock Efficient video diffusion models via content-frame motion-latent decomposition.
\newblock {\em arXiv preprint arXiv:2403.14148}, 2024.

\bibitem{ControlNet}
Lvmin Zhang and Anyi Rao.
\newblock Adding conditional control to text-to-image diffusion models.
\newblock In {\em ICCV}, 2023.

\bibitem{zhang2018lpips}
Richard Zhang and Isola.
\newblock The unreasonable effectiveness of deep features as a perceptual metric.
\newblock In {\em CVPR}, 2018.

\bibitem{sadtalker2023}
Wenxuan Zhang and Xiaodong Cun.
\newblock Sadtalker: Learning realistic 3d motion coefficients for stylized audio-driven single image talking face animation.
\newblock In {\em CVPR}, 2023.

\bibitem{zhang2024musetalk}
Yue Zhang, Minhao Liu, and Zhaokang Chen.
\newblock Musetalk: Real-time high quality lip synchronization with latent space inpainting.
\newblock {\em arXiv preprint arXiv:2410.10122}, 2024.

\bibitem{hdtf2021}
Zhimeng Zhang, Lincheng Li, and Yu~Ding.
\newblock Flow-guided one-shot talking face generation with a high-resolution audio-visual dataset.
\newblock In {\em CVPR}, 2021.

\bibitem{zhao2022thin}
Jian Zhao and Hui Zhang.
\newblock Thin-plate spline motion model for image animation.
\newblock In {\em CVPR}, 2022.

\bibitem{zhou2024storydiffusion}
Yupeng Zhou and Zhou.
\newblock Storydiffusion: Consistent self-attention for long-range image and video generation.
\newblock In {\em NeurIPS}, 2024.

\end{thebibliography}
